\documentclass[lettersize,journal]{IEEEtran}
\usepackage{amsmath,amsfonts}
\usepackage{algorithmic}
\usepackage{algorithm}
\usepackage{array}
\usepackage[caption=false,font=normalsize,labelfont=sf,textfont=sf]{subfig}
\usepackage{array}     
\usepackage{booktabs}  
\usepackage{multirow}  
\usepackage{color}
\usepackage{textcomp}
\usepackage{stfloats}
\usepackage{url}
\usepackage{verbatim}
\usepackage{graphicx}
\usepackage{xcolor}
\usepackage{cite}

\usepackage[colorlinks,
            linkcolor=green,
            anchorcolor=green,
            citecolor=green]{hyperref}

\begin{document}

\title{PhysioSync: Temporal and Cross-Modal Contrastive Learning Inspired by Physiological Synchronization for EEG-Based Emotion Recognition}
% \title{PhysioSync: Physiological Synchronization Contrastive Learning and Cross-Modal Consistency Alignment for EEG-based Emotion Recognition}
% \title{TCL-CCA: Temporal Contrastive Learning and Cross-Modal Consistency Alignment for EEG-based Emotion Recognition}
% \title{Contrastive Learning of Affective Stimulus Alignment for Multimodal physiology signals Emotion Recognition}

\author{Kai Cui, Jia Li, Yu Liu, Xuesong Zhang, Zhenzhen Hu,  Meng Wang, \IEEEmembership{Fellow, IEEE}
\thanks{
This work was supported in part by the National Natural Science Foundation of China under Grants U23A20294 and 62202139, and in part by the Fundamental Research Funds for the Central Universities under Grant JZ2025HGTB0226. \textit{(Kai Cui and Jia Li contributed equally to this work.) (Corresponding authors: Jia Li, Yu Liu.)}}
\thanks{
Kai Cui, Yu Liu are with the School of Instrument Science and Opto-electronics Engineering, Hefei University of Technology, Hefei 230009, China (e-mail: kaic@mail.hfut.edu.cn; yuliu@hfut.edu.cn).}
\thanks{
Jia Li, Xuesong Zhang, Zhenzhen Hu, Meng Wang are with the School of Computer Science and Information Engineering, Hefei University of Technology, Hefei 230009, China (e-mail: jiali@hfut.edu.cn; xszhang\_hfut@mail.hfut.edu.cn; 
zzhu@hfut.edu.cn; eric.mengwang@gmail.com).
% (\emph{Jia Li and Kai Cui contributed equally to this work}.) 
\\
% (\emph{Corresponding author: Yu Liu, Meng Wang}.)
}} 
\maketitle

\begin{abstract}

%% lijia2
Electroencephalography (EEG) signals provide a promising and involuntary reflection of brain activity related to emotional states, offering significant advantages over behavioral cues like facial expressions.
However, EEG signals are often noisy, affected by artifacts, and vary across individuals, complicating emotion recognition. While multimodal approaches have used Peripheral Physiological Signals (PPS) like GSR to complement EEG, they often overlook the dynamic synchronization and consistent semantics between the modalities. Additionally, the temporal dynamics of emotional fluctuations across different time resolutions in PPS remain underexplored.
To address these challenges, we propose PhysioSync, a novel pre-training framework leveraging temporal and cross-modal contrastive learning, inspired by physiological synchronization phenomena. PhysioSync incorporates Cross-Modal Consistency Alignment (CM-CA) to model dynamic relationships between EEG and complementary PPS, enabling emotion-related synchronizations across modalities. Besides, it introduces Long- and Short-Term Temporal Contrastive Learning (LS-TCL)  to capture emotional synchronization at different temporal resolutions within modalities.
After pre-training, cross-resolution and cross-modal features are hierarchically fused and fine-tuned to enhance emotion recognition.
Experiments on DEAP and DREAMER datasets demonstrate PhysioSync’s advanced performance under uni-modal and cross-modal conditions, highlighting its effectiveness for EEG-centered emotion recognition. The source code will be publicly available at \url{https://github.com/MSA-LMC/PhysioSync}.

\end{abstract}

\begin{IEEEkeywords}
EEG-based emotion recognition, multi-modal fusion,  contrastive-learning.
\end{IEEEkeywords}

\section{Introduction}

%% zxs3

\IEEEPARstart{R}{ecognizing} emotional states through physiological signals is a critical yet challenging task in human-computer interaction, disease diagnosis, and rehabilitation, attracting substantial research attention recently~\cite{zhang2023discriminative,jafari2023emotion,alarcao2017emotions,mauss2009measures}.  Compared to facial expressions, EEG stands out as a non-invasive, cost-effective modality with high temporal resolution, offering direct and involuntary reflections of emotional states, thus establishing itself as the cornerstone of emotion recognition research~\cite{li2022sstd,li2024gusa, ye2024adaptive}.  
Despite its advantages, EEG-based emotion recognition faces challenges such as overfitting to noise from physiological artifacts and insensitivity to inter-subject variability \cite{gu2022multi}, which recent studies~\cite{tang2024hierarchical,shen2024tensor,zhu2023dynamic,fu2024cross} suggest can be mitigated by incorporating Peripheral Physiological Signals (PPS) to provide complementary information.
Notably, some studies suggest that EEG is better for predicting arousal, while PPS are more suited for assessing valence\cite{arthanarisamy2022subject}, highlighting the benefit of integrating both.
Physiological synchronization, which denotes the simultaneous temporal alignment of multiple physiological signals elicited by the same emotional stimulus, captures coordinated bodily responses and plays a key role in enhancing multimodal emotion recognition\cite{ hasson2004intersubject, stuldreher2020physiological, zhang2023multimodal, kleinbub2020physiological}.
From this perspective, simply incorporating complementary modalities does not sufficiently address the challenge of cross-subject variability and may even exacerbate modality heterogeneity, posing a significant obstacle to the advancement of multimodal emotion recognition.

\begin{figure}[!t]
    \centering
    \includegraphics[width=0.48\textwidth]{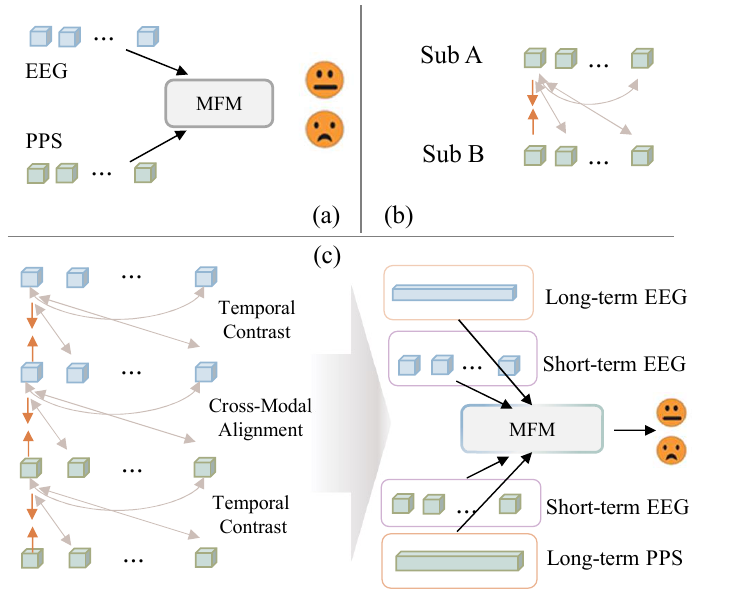}
    \caption{Overview of the proposed approach compared to existing methods. (a) represents the simple combination of EEG and peripheral physiological signals (PPS) (w/o pre-training); (b) involves contrastive learning (w/ pre-training); (c) represents our method (w/ pre-training). Orange arrows denote positive pairs, while gray arrows indicate negative pairs. MFM refers to the modality fusion module.}
    \label{fig:models}
\end{figure}

To address these challenges, current research efforts are primarily focused on multi-modality heterogeneity and cross-subject divergence.
Some approaches combine EEG with Peripheral Physiological Signals (PPS) (Fig. \ref{fig:models} (a)). For example, Zhu et al.~\cite{zhu2023dynamic} introduced facial expressions in conjunction with EEG signals for dynamic, confidence-aware multimodal fusion and prediction. Similarly, Fu et al.~\cite{fu2024cross} proposed a cross-modal guiding neural network, utilizing EEG features to guide the extraction of eye movement features, thereby mitigating the impact of subjective factors. These studies underscore the value of PPS as supplementary information to EEG. 
However, there is a notable gap in research concerning consistent semantic alignment between different modalities and EEG signals to overcome multi-modal heterogeneity. 
For effective multimodal emotion recognition, it is essential that signals from different modalities are semantically aligned\cite{li2021align}. Intuitively, signals from different modalities should be correlated when exposed to the same emotional stimuli.
In addition to multi-modal heterogeneity, cross-subject variability continues to be a pervasive challenge. To tackle this, recent studies have employed contrastive learning to learn spatiotemporal representations that align EEG signals across subjects (Fig. \ref{fig:models} (b)). 
For example, Shen et al.\cite{shen2022contrastive} utilized convolutional neural networks to align EEG features across subjects using contrastive learning. 
However, these methods typically apply contrastive learning at a single temporal resolution, while emotional fluctuations inherently involve both short-term and long-term correlations\cite{houben2015relation}.

In this paper, we tackle the heterogeneity between complementary PPS and EEG signals, as well as cross-subject variability, through cross-modal alignment and contrastive learning. Additionally, we capture the dynamic characteristics of emotion by extracting and integrating features across multiple temporal scales. Specifically, we propose a contrastive learning framework for physiological synchronization that integrates EEG with PPS (Fig. \ref{fig:models} (c)).
%A data augmentation step is employed to generate more training samples and improve emotion-related semantic consistency learning against moderate noise. 
Our contrastive learning framework includes two key components: To minimize discrepancies in physiological signals induced by the same emotional stimulus across subjects, we propose TCL (intra-modal temporal contrastive learning). To align signals from different modalities in a shared semantic space, we introduce CM-CL (cross-modal contrastive learning). 
This dual branch facilitates intra-modality temporal synchronization for feature extraction, alongside inter-modality synchronization and interaction. 
Encoders that extract long-term and short-term features are pre-trained separately based on the different resolutions of the input signal.
In the fine-tuning stage, long-term physiological signals are decomposed into short-term segments, which are then processed by the pre-trained encoders to extract both long-term and short-term features, capturing the temporal dynamics of emotional states and offering a comprehensive perspective for emotion recognition.

Specifically, our contributions are as follows:

\begin{itemize}
\item A temporal and cross-modal contrastive learning framework, inspired by physiological synchronization phenomena, is propose to pre-train emotional representations in time domain. It aligns features intra- and inter-modal within same stimulus segments, capturing consistent emotional patterns across physiological signals. %subjects
\item We pre-trains representations on different temporal windows (e.g., 1-second and 5-second windows) for EEG and PPS, effectively modeling short- and long-term emotional dynamics. During fine-tuning, fusing features across temporal resolutions and modalities significantly improves  emotion recognition accuracy.
\item Our method directly processes raw EEG and PPS as temporal sequences using a Transformer-based backbone, eliminating the need for handcrafted features like differential entropy, and providing a simple, reproducible baseline for future research in EEG-based emotion recognition.
\end{itemize}

\section{RELATED WORK}
\subsection{EEG-Based Multimodal Emotion Recognition}
Electroencephalographic (EEG) signals, reflecting brain activity and psychological processes, contain vital psycho-physiological information\cite{alarcao2017emotions}. Unlike other physiological signals, EEG is directly tied to the central nervous system, offering more precise emotional insights \cite{liu2024eeg}. This advantage has driven the development of various EEG-based emotion recognition methods \cite{liu2024capsnet}, \cite{du2020efficient}. However, many methods still rely on hand-crafted features like power spectral density (PSD) and differential entropy (DE), rather than fully exploiting the potential of deep learning models.
In contrast, some approaches use raw EEG data with advanced techniques like CNNs and LSTMs to capture spatial and temporal features. For instance, Yin et al. \cite{yin2021eeg} proposed a framework combining graph CNNs and LSTMs, while Ding et al. \cite{ding2022tsception} introduced TSception, a multi-scale CNN that integrates dynamic temporal, asymmetric spatial, and high-level fusion layers for more effective representation learning across temporal and channel dimensions.

Despite these advancements, the complexity of human emotional states poses a significant challenge. A single EEG modality is insufficient to provide a comprehensive representation of the current emotional state, and achieving satisfactory recognition in terms of both accuracy and robustness remains a challenge \cite{he2020advances}. 
To address this, researchers have explored methods that combine signals from the central nervous system \cite{kwak2022fganet} (e.g., EEG) and the peripheral nervous system \cite{zitouni2022lstm} (e.g., GSR, ECG, EMG, EOG, etc.). 
Jimenez et al.\cite{jimenez2024mmda} proposed a multi-modal and multi-source Domain Adaptation (MMDA) method for addressing the multi-modal emotion recognition problem using EEG and eye movement signals. Li et al.\cite{li2024incongruity}  presented a fusion model that leverages the Cross Modal Transformer (CMT), Low Rank Fusion (LRF), and modified CMT (MCMT) to reduce incongruity and redundancy among multimodal physiological signals.

\subsection{Contrastive Learning for Physiological Representation}

Contrastive learning is a form of self-supervised learning that aims to learn representations without relying on manual labels. It has demonstrated state-of-the-art performance across various domains, including computer vision (CV) \cite{chen2020simple}, natural language processing (NLP) \cite{qu2020coda}, and bioinformatics \cite{li2022supervised}. Typically, contrastive learning methods involve pre-training representations on relatively large datasets, which are then fine-tuned for downstream tasks \cite{liu2023self}. However, in the domain of physiological signal emotion recognition, particularly in Electroencephalogram (EEG)-based emotion recognition, the absence of large-scale datasets presents a significant challenge for pre-training. To address this, Shen et al. \cite{shen2022contrastive} proposed a contrastive learning strategy tailored for cross-subject generalization. This approach learns the similarities between samples from different subjects exposed to the same stimuli, enabling the model to generalize across subjects. Notably, their method does not rely on large external datasets; instead, it generates numerous self-supervised labels based on the alignment of experimental designs across subjects.

One form of contrastive learning involves classifying samples into positive and negative pairs based on the internal relationships within the data. Using a specific loss function, this approach aims to maximize the similarity between positive pairs while minimizing the similarity between negative pairs \cite{chen2020simple}. Kan et al.\cite{kan2023self} utilized self-supervised learning alongside a novel genetics-inspired data augmentation method to achieve state-of-the-art performance in EEG-based emotion recognition. Simultaneously, contrastive learning methods have also been introduced in cross-corpus EEG-based emotion recognition. Liu et al.\cite{liu2024joint} introduced a cross-domain contrastive learning strategy during the pre-training phase and incorporating inter-electrode structural connectivity during the fine-tuning phase, the method significantly improved the accuracy of emotion recognition across different datasets. Our work introduces a multi-modal contrastive learning approach that effectively integrates both temporal and cross-modal contrastive learning to better combine multi-modal information.

\begin{figure*}[!tb]
    \centering
    \includegraphics[width=0.93\textwidth, height=0.37\textwidth]{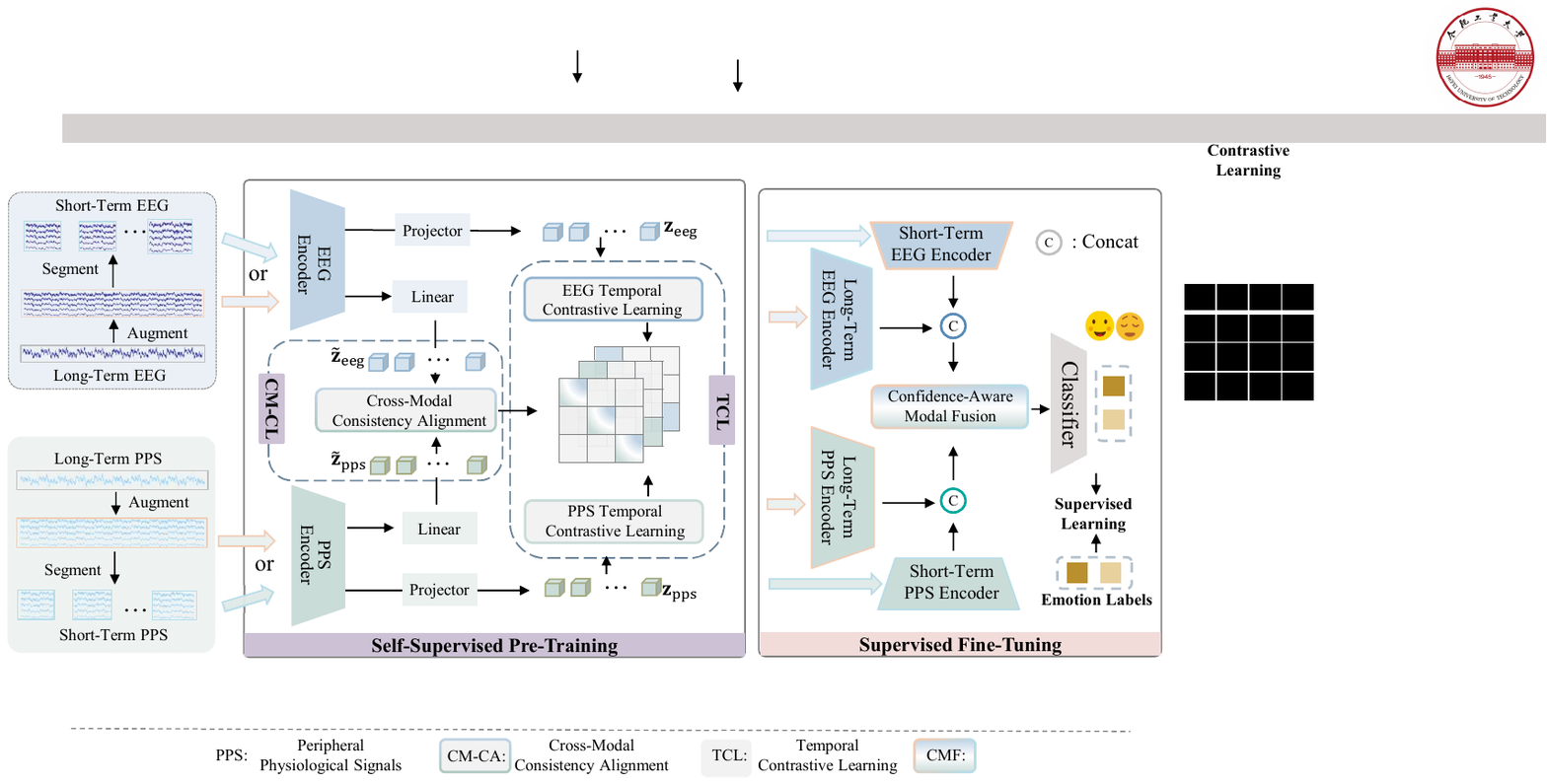}
    \caption{The proposed contrastive learning framework inspired by physiological synchronization, which consists of a self-supervised pre-training phase and a supervised fine-tuning phase. CM-CL denotes the Cross-Modal Contrastive Learning, while TCL refers to the Temporal Contrastive Learning. Fig. \ref{fig:structure1} provides the details of the Encoder and Projector. Fig. \ref{fig:structure2} illustrates the Fusion Module and Classifier.}
    \label{fig:model1}
\end{figure*}

\section{PROPOSED METHOD}
This section introduces PhysioSync, a novel contrastive learning framework with two phases: self-supervised pre-training and supervised fine-tuning (Fig. \ref{fig:model1}). Section A covers the data input and augmentation for multimodal physiological signals. Section B presents the temporal and cross-modal contrastive optimization mechanism. Section C describes the fine-tuning strategy, aligning pre-trained features with emotion labels through task-specific layers.

\subsection{Preliminaries}
% \subsubsection{Problem Formulation}
    
\subsubsection{Long- and Short-term Clip Segmentation}
To capture multi-scale emotional features, we set different time lengths of \(t\) to train encoders for long-term and short-term feature extraction, respectively.
Specifically, the input is processed in mini-batches from subjects A and B as an example, the data from each trial of subject A is divided into \(\boldsymbol{t}\)-second clips in temporal order, resulting in \(N\) clips denoted as \(\boldsymbol{X}^{\text{eeg}}_{i,A}\) (\(i = 1, 2, 3, \ldots, N\)), where \(\boldsymbol{X}^{\text{eeg}}_{i,A} \in \mathbb{R}^{C \times S}\), \(C\) is the number of channels, and \(S\) is the number of sampling points. Similarly, subject B's eeg data is divided into \(\boldsymbol{X}^{\text{eeg}}_{i,B}\) (\(i = 1, 2, 3, \ldots, N\)). segments \(\boldsymbol{X}^{\text{eeg}}_{i,A}\) and \(\boldsymbol{X}^{\text{eeg}}_{i,B}\) from the same video stimulus form a positive pair, while \(\boldsymbol{X}^{\text{eeg}}_{i,S}\) and \(\boldsymbol{X}^{\text{eeg}}_{j,S}\) (\(j \neq i, S \in \{A, B\}\)) from different video stimuli form a negative pair. For all subjects and other modalities, such segmentation is performed, then segments for \(K\) video stimuli from 2 subjects are randomly sampled to form a mini-batch \(\boldsymbol{G}_{m} = \{\boldsymbol{X}^{\text{m}}_{i,S} \mid i = 1, 2, 3, \ldots, K; S \in \{A, B\}\}\) as input for pre-training. Here, \(m\) can represent both eeg or pps.

\subsubsection{ Data Augmentation}
A major challenge in using physiological signals for emotion recognition is the limited number of subjects and small public datasets, which can hinder model training \cite{singh2023subject}. To address this, data augmentation techniques, such as scaling and noise addition, can be employed to expand the dataset \cite{lopez2023hypercomplex}, and introduce more positive samples for contrastive learning. The data augmentation step generates more training samples and improves emotion-related semantic consistency learning against moderate noise. Specifically, two scaling factors are applied to the original samples, and Gaussian noise with a fixed mean and signal-to-noise ratio (SNR) is added. The process can be summarized as follows:
\begin{equation}
\tilde{\boldsymbol{G}_m} = \mathcal{A}(G_m),
\end{equation}
where \(\mathcal{A}\) represents the data augmentation operation. Through scaling, noise addition, and their combination, the data is augmented to five times its original size. Finally, the input can be represented as: 
\begin{equation}
\tilde{\boldsymbol{G}_m} = \{\boldsymbol{\tilde{X}}^{m}_{i,S} \,|\, i = 1, 2, 3, \ldots, 5K ; S \in \{A, B\}\}.
\end{equation}

\subsection{Self-supervised Pre-training Strategy}
\subsubsection{Pretraining model}
\begin{figure}[tb] % figure for single column
    \centering
    \includegraphics[width=0.33\textheight,height=0.26\textheight]{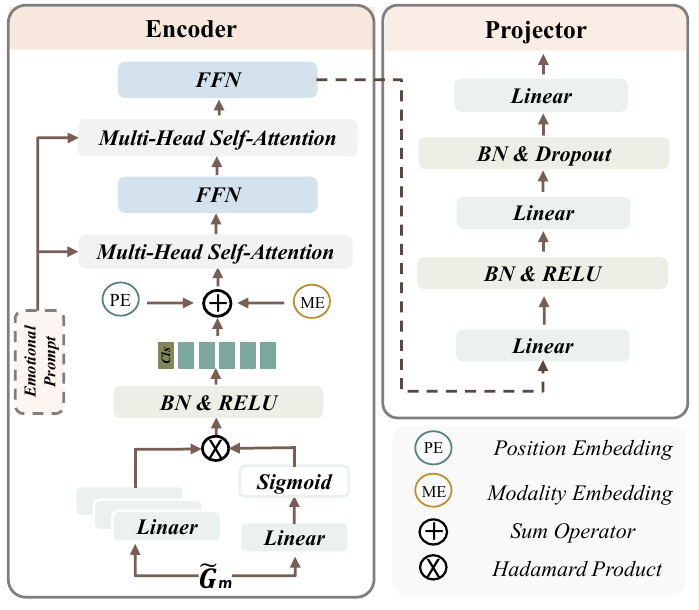} 
    \caption{The illustration of Encoder and Projector, where the Encoder we take part of the structure from\cite{jiang2023multimodal}.}
    \label{fig:structure1}
\end{figure}

Previous work has shown that self-supervised pre-training strategies can significantly aid model learning. Building upon the structure from \cite{jiang2023multimodal}, we have designed an encoder, as illustrated in Fig. \ref{fig:structure1}.

Physiological signals are input and transformed into \(\boldsymbol{z}\) embeddings through parallel linear layers, aiming to encourage the model to focus on different views of features. Another linear layer followed by an activation function is used to gate the embeddings with information useful for emotion recognition. Subsequently, \( \boldsymbol{E}_i \) and \( \boldsymbol{\tilde{E}} \) are multiplied element-wise and aggregated across the \(\boldsymbol{z}\) embeddings to obtain \( \boldsymbol{e} = (\boldsymbol{e}_1, \dots, \boldsymbol{e}_z) \). The formula is as follows \cite{jiang2023multimodal}:

\begin{equation}
\boldsymbol{E}_i = \textit{Linear}_i(\tilde{\boldsymbol{G}}_m), \quad i = 1, \ldots, z,
\end{equation}
\begin{equation}
\boldsymbol{\tilde{E}} = \sigma (\textit{Linear}(\tilde{\boldsymbol{G}}_m)),
\end{equation}
\begin{equation}
\boldsymbol{e}_i = \textit{RELU}(\textit{BN}(\textit{Stack}(\boldsymbol{\tilde{E}}  \odot \boldsymbol{E}_i))), \quad i = 1, \ldots, z,
\end{equation}
where \(\boldsymbol{\sigma}\) represents the sigmoid function, which constrains the output values between 0 and 1, and \(\boldsymbol{\odot}\) denotes the Hadamard product. BN stands for batch normalization, and RELU denotes the activation function. By this method, the input features \(\tilde{\boldsymbol{G}}_m\)\ are transformed into token sequences from different views, which can be further processed by subsequent layers.

Before passing to the next layers, the embedding \(\boldsymbol{e}\) is prepended with a learnable class token \( \boldsymbol{E}_{cls} \), which serves to aggregate information from the entire sequence. To integrate positional and modality information, learnable positional embeddings \( \boldsymbol{E}_{pos} \) and modality embeddings \( \boldsymbol{E}_{mod} \) are added to the input embeddings. This approach ensures that the input encompasses not only the sequence content but also positional and modality information:
\begin{equation}
\boldsymbol{\tilde{e}} = [\boldsymbol{E}_{\textit{cls}}, \boldsymbol{e}_1, \ldots, \boldsymbol{e}_z] + \boldsymbol{E}_{\textit{pos}} + \boldsymbol{E}_{\textit{mod}}.
\end{equation}

The core component of the Transformer is the multi-head self-attention mechanism \cite{vaswani2017attention}. The embedding \( \boldsymbol{\tilde{e}} \) is transformed into query \(\boldsymbol{Q}\), key \(\boldsymbol{K}\), and value \(\boldsymbol{V}\) through three linear layers, with the calculation formula as follows:
\begin{equation}
\textit{Attention}(\boldsymbol{Q}, \boldsymbol{K}, \boldsymbol{V}) = \textit{softmax}\left(\frac{\boldsymbol{Q} \boldsymbol{K}^T}{\sqrt{\boldsymbol{d_e}}}\right) \boldsymbol{V},
\end{equation}
where \(\boldsymbol{d_e}\) is the dimension of the embedding. Then we use \(\boldsymbol{h}\) heads for self-attention, with each head represented as \(\boldsymbol{H} = \textit{Attention}(\boldsymbol{Q}, \boldsymbol{K}, \boldsymbol{V})\). The output of the multi-head attention is \(\textit{Concat}(\boldsymbol{H}_1, \boldsymbol{H}_2, \ldots, \boldsymbol{H}_h)\boldsymbol{W}\), where \(\boldsymbol{W}\) is the weight matrix. Subsequently, after further processing by a feed-forward neural network (FFN), then repeat twice. Incorporating emotional prompt (learnable token) \cite{darcet2023vision} boosts the model's sensitivity to emotional changes. We can simply 
denote the entire encoder as ${f_m}$:
\begin{equation}
\boldsymbol{H}_{i,S}^m = {f_m}(\boldsymbol{\tilde{X}}_{i,S}^m).
\end{equation}

\subsubsection{Long- and Short-Term Temporal Contrastive Learning}
To align the stimuli across subjects and achieve physiological synchronization among different subjects, the features extracted by the encoder are used for intra-modal temporal contrastive learning (TCL). Different values of \(t\) are used to perform contrastive learning at different time resolutions, and the corresponding encoders are pre-trained accordingly. Inspired by the SimCLR framework \cite{chen2020simple}, a nonlinear projection head is employed between the encoder and the final contrastive loss. The structure, shown in Fig. \ref{fig:structure1}, is used to process the features extracted by the encoder:
\begin{equation}
\begin{array}{c}
\boldsymbol{h}_{1,i,S}^m = \mathit{ReLU}\left( \mathit{BN}(W_1 \boldsymbol{H}_{i,S}^m + b_1) \right) \\
\boldsymbol{h}_{2,i,S}^m = \mathit{Dropout}\left( \mathit{BN}(W_2 \boldsymbol{h}_{1,i,S}^m + b_2) \right) \\
\boldsymbol{z}_{i,S}^m = W_3 \boldsymbol{h}_{2,i,S}^m + b_3
\end{array}
,
\end{equation}
where \(W_1, W_2, W_3 \) are the weight matrices of the three linear layers; \(b_1, b_2, b_3 \) are the bias vectors and \(\boldsymbol{h}_{2,i,S}^m, \boldsymbol{h}_{1,i,S}^m, \boldsymbol{z}_{i,S}^m \) denote the output features of each layer. where \(i\) denotes the number of physiological signal segments, \(S\) denotes the subject. Subsequently, the similarity between samples is calculated using the following formula:
\begin{equation}
s(\boldsymbol{z}_{i,A}^m, \boldsymbol{z}_{j,B}^m) = \frac{\boldsymbol{z}_{i,A}^m \cdot \boldsymbol{z}_{j,B}^m}{\|\boldsymbol{z}_{i,A}^m\|\|\boldsymbol{z}_{j,B}^m\|}, \quad s(\boldsymbol{z}_{i,A}^m, \boldsymbol{z}_{j,B}^m) \in [0, 1]
\end{equation}

Similar to the SimCLR, we use the normalized temperature-scaled cross-entropy to define the loss function as follows:
\begin{equation}
\begin{array}{c}
\mathcal{L}_{i, A}^m = -\mathit{log}\left( \frac{\mathit{S}_1}{\mathit{S}_2 + \mathit{S}_3} \right) \\
\mathit{S}_1 = \mathit{exp}\left( \frac{s(\boldsymbol{z}_{i, A}^m, \boldsymbol{z}_{i, B}^m)}{\tau} \right) \\
\mathit{S}_2 = \sum_{j=1}^{5K} \mathbb{I}_{[j \neq i]}  \mathit{exp}\left( \frac{s(\boldsymbol{z}_{i, A}^m, \boldsymbol{z}_{j, A}^m)}{\tau} \right) \\
\mathit{S}_3 = \sum_{j=1}^{5K} \mathit{exp}\left( \frac{s(\boldsymbol{z}_{i, A}^m, \boldsymbol{z}_{j, B}^m)}{\tau} \right)
\end{array}
,
\end{equation}
where the indicator function \(\mathbb{I}_{[i \neq j]}\) equals 1 when \(i \neq j\), and 0 otherwise. By minimizing this loss function, the model increases the similarity of positive pairs while reducing the similarity of negative pairs. 
The loss for the mini-batch \(L_m\) is given by: 
\begin{equation}
L_m = \sum_{i=1}^{5K} (\mathcal{L}_{i,A}^m + \mathcal{L}_{i,B}^m).
\end{equation}

\subsubsection{Cross-Modal Consistency Alignment}
To align physiological signals from different modalities, the features extracted by the encoder are also used for cross-modal contrastive learning (CM-CL). Physiological signal segments from different modalities induced by the same stimulus are treated as positive pairs, while others are treated as negative pairs. Additionally, before calculating the similarity, the features pass through a linear layer to obtain \(\boldsymbol{\tilde{z}}_{i,s}^m\). Then according to formulas (10)–(12), the cross-modal contrastive loss \(L_{\text{cc}}\) is computed.

During the pre-training phase, the total loss \( L \) consists of the contrastive loss within the EEG modality \( L_{eeg} \), the contrastive loss within peripheral physiological signals \( L_{pps} \), and the cross-modal contrastive loss \( L_{cc} \):
\begin{equation}
L = \alpha \cdot L_{eeg} + \beta \cdot L_{pps} + \gamma \cdot L_{cc}.
\end{equation}

We empirically set \(\alpha = 0.5\), \(\beta = 0.5\) to balance EEG and PPS supervision, and \(\gamma = 1\) to emphasize cross-modal consistency and enhance feature alignment.

After pre-training, the encoder can effectively extract consistency features across different modalities at various time resolutions. Algorithm \ref{alg:clasa_pretraining} summarizes the pre-training process.

\begin{algorithm}[t]
\caption{The Pre-Training Procedure for PhysioSync}
\label{alg:clasa_pretraining}
\begin{algorithmic}[1]
\STATE \textbf{Input:} Training data $\boldsymbol{X}^{eeg}$ and $\boldsymbol{X}^{pps}$
\STATE Initialize the parameters of the base encoders $\theta^{eeg}$, $\theta^{pps}$ and the projector $\theta^{pr\_eeg}$, $\theta^{pr\_pps}$
\FOR{epoch = 1 to $T$}
    \REPEAT
        \STATE Sample two subjects $A$, $B$, Random scramble after segmentation $\{\boldsymbol{X}_{i,S}^{m} | i=1, \ldots, N; S \in \{A, B\}\}$
        \STATE Randomly select $K$ segments $\boldsymbol{G}_m = \{\boldsymbol{X}_{i,S}^{m} | i=1, \ldots, K; S \in \{A, B\}\}$
        \STATE Obtain $\widetilde{\boldsymbol{G}}_m = \{\boldsymbol{\widetilde{X}}_{i,S}^{m} | i=1, \ldots, 5K; S \in \{A, B\}\}$ by data augmentation
        \STATE Obtain $\{\boldsymbol{z}_{i,S}^{m} | i=1, \ldots, 5K; S \in \{A, B\}\}$ by (3)-(9)
        \STATE Calculate loss $L_m$ and $L_{cc}$ by (10)-(12), then obtain total loss $L$ by (13)
        \STATE Abate loss $L$ through optimizer updating parameters of $\theta^{eeg}$, $\theta^{pps}$, $\theta^{pr\_eeg}$ and $\theta^{pr\_pps}$
    \UNTIL{all possible pairs of subjects are enumerated}
\ENDFOR
\STATE \textbf{Output:} Parameters $\theta^{eeg}$, $\theta^{pps}$
\end{algorithmic}
\textit{Where $T$ is the training epochs.}
\end{algorithm}

\subsection{Supervised Fine-tuning Framework}
% \subsubsection{Long- and Short Term Feature Fusion}  
During the pre-training stage, we obtain modality-specific encoders at different time resolutions, which are subsequently used for supervised fine-tuning in the downstream task. Specifically, the pre-trained encoders are employed to extract consistency features for EEG and PPS at long or short time resolutions (the short-term input is derived by decomposing the long-term input into \( 1s \) ). Then their long-term and short-term features are fused by concatenation separately into \( \boldsymbol{H}_{eeg} \) and \( \boldsymbol{H}_{pps} \), which are then input into the modal fusion module.

\begin{figure}[t] % figure for single column
    \centering
    \includegraphics[width=0.33\textheight,height=0.26\textheight]{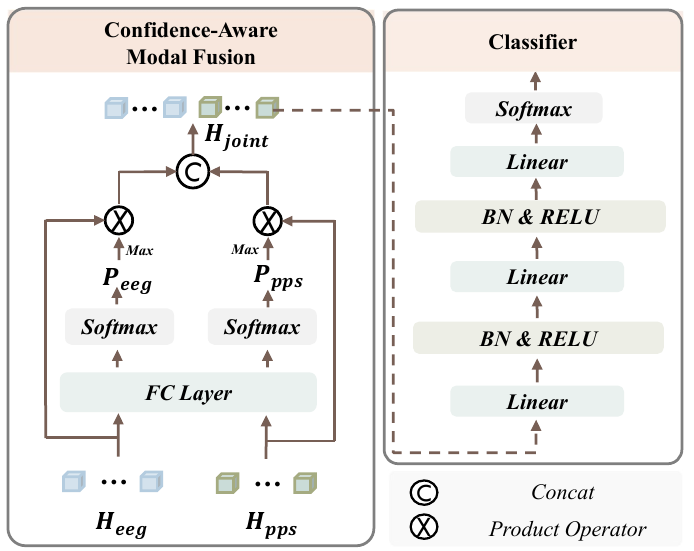} % replace example-image with your image file name
    \caption{The illustration of Modal Fusion Module and Classifier. In the Modal Fusion Module, we employ the Maximum Class Probability (MCP) to calculate confidence levels and assign weights to each modality accordingly.}
    \label{fig:structure2}
\end{figure}

% \subsubsection{Confidence-Aware Multi-Modal Fusion}
In the Modal Fusion Module (Fig. \ref{fig:structure2}), to flexibly adjust the information flow, we use Maximum Class Probability (MCP) \cite{zhu2023dynamic} to compute the confidence level of each modality and assign weights to the features based on their confidence. MCP enhances the robustness and accuracy of the fusion process by measuring the confidence of each modality, ensuring more reliable decision-making in the presence of uncertain or noisy data:
\begin{equation}
\text{MCP}(\boldsymbol{H}_m) = \max_{k \in \boldsymbol{y}} P(Y = k \mid w, \boldsymbol{H}_m),
%= P(Y = \boldsymbol{\hat{y}} \mid w, \boldsymbol{H}_m)
%and \( \boldsymbol{\hat{y}} \) corresponds to the predicted class for the sample,
\end{equation}
where \( w \) represents the set of network parameters,  \(m\) can represent both eeg or pps. Subsequently, the weighted features from both modalities are concatenated. This combined feature $\boldsymbol{H}_{joint}$ is then fed into a classifier for final classification, and we optimize the model using cross-entropy loss.
\begin{equation}
L_{CE} = -\boldsymbol{y} \log P(\boldsymbol{H}_{joint}) - (1 - \boldsymbol{y}) \log (1 - P(\boldsymbol{H}_{joint})),
\end{equation}
where \( P(\boldsymbol{H}_{joint}) \) represents the probability that the joint feature vector belongs to a certain class, and \( \boldsymbol{y} \) denotes the sample label.

\section{EXPERIMENTS}
Our method PhysioSync is experimented and evaluated on two widely used multimodal physiological signaling datasets: the DEAP dataset and the DREAMER \cite{koelstra2011deap},\cite{katsigiannis2017dreamer}. This section first describes the DEAP and DREAMER datasets in detail and the related implementation details. We then show the results of various experiments on both datasets.

\subsection{Dataset}
\subsubsection{DEAP}
A multimodal physiological signal dataset used for emotion recognition, encompassing electroencephalogram (EEG) signals and several peripheral physiological signals. We primarily utilized EEG, galvanic skin response (GSR), electromyography (EMG), and electrooculography (EOG). These signals were recorded from 32 subjects as they watched 40 different short videos, with each recording session lasting 63 seconds (60 seconds for emotional arousal and 3 seconds for baseline). The Self-Assessment Manikin (SAM)\cite{bradley1994measuring} was utilized to capture the emotional ratings of the subjects, which included assessments of arousal, valence, liking, and dominance for each video, rated on a scale from 1 to 9. In this experiment, we utilized arousal and valence as the classification criteria. When the ratings are greater than or equal to 5, they are labeled as high arousal or high valence. Conversely, ratings below 5 are labeled as low arousal or low valence. This approach forms a binary classification for each dimension, and the combination of the two dimensions results in a four-class classification.
\subsubsection{DREAMER}
The dataset comprises 14-channel EEG and 2-channel Electrocardiogram (ECG) data from 23 subjects (14 males and 9 females). Each subject viewed 18 movie clips, with durations ranging from 65 to 393 seconds. These clips were designed to elicit various emotions, including amusement, excitement, happiness, calmness, anger, disgust, fear, sadness, and surprise. Following each clip, subjects provided subjective evaluations of valence, arousal, and dominance using the Self-Assessment Manikin (SAM) on a scale from 1 to 5. The midpoint of the scale served as the threshold value: ratings of 3 or below indicated low valence or arousal, while ratings above 3 indicated high valence or arousal.

% \begin{table}[h!]
%     \centering
%     \caption{Summary of the DEAP, DREAMER databases.}
%     \label{table:dataset}
%     \scalebox{1.1}{%
%         \begin{tabular}{lcc}
%             \toprule
%             \textbf{Databases}              & \textbf{DEAP} & \textbf{DREAMER} \\ 
%             \midrule
%             Participants                   & 32            & 23              \\ 
%             Emotional stimuli              & Musical videos & Movie clips     \\ 
%             Trials for a participant       & 40            & 18              \\ 
%             Physiological modalities       & 7             & 2               \\ 
%             Target emotional classes       & Binary        & Binary          \\ 
%             \bottomrule
%         \end{tabular}
%     }
   
% \end{table}

\subsection{Data Pre-processing}
In the DEAP dataset, we used the publicly available preprocessed signals, which include downsampling to 128 Hz, EOG artifact removal \cite{koelstra2011deap}, 4–45 Hz bandpass filtering, common average re-referencing, and EEG channel reordering following the Geneva scheme. Each trial contains a 60-second emotion segment and a 3-second baseline.  Following \cite{yang2018emotion}, we averaged the baseline into a 1-second reference and subtracted it from the emotion segment. The corrected signals were segmented into 5-second clips, yielding 480 clips per participant. With 32 participants and 40 videos, this resulted in 15,360 segments for 10-fold cross-validation.

For the DREAMER dataset, we used the raw data provided by the official release. No artifact removal or filtering was applied, except that the ECG signals were downsampled to 128 Hz to match the sampling rate of other modalities. A similar baseline correction procedure was applied. The data was then segmented into 1-second windows for both pre-training and classification stages.

During the pre-training phase, we organized the dataset tensor along five dimensions: \textit{video clips}, \textit{subjects}, \textit{1}, \textit{channels}, and \textit{sampling points}. In each fold, every epoch iterated through all the \textit{video clips} in the current training data. In the fine-tuning phase, applying the same criteria, we divided the data along the \textit{video clips} dimension and then reshaped the first two axes—\textit{video clips} and \textit{subjects}—into a sample axis. After reshaping, each axis of the dataset corresponded to \textit{samples}, \textit{1}, \textit{channels}, and \textit{sampling points}.

\begin{table*}[!htb]
    \centering
    \begin{minipage}{\textwidth}
        \centering % 表格和描述一起居中
        \caption{Performance comparison with SOAT schemes in terms of accuracy (\%) and F1 score (\%) on the DEAP dataset. Where * indicates that the results of this model were reproduced by ourselves, while the others are results reported in the paper.}
        \label{tab:deap_sota}
        \setlength{\tabcolsep}{0.1pt} 
        \scalebox{0.83}{%
        \begin{tabular}{@{}c c*{6}{>{\centering\arraybackslash}p{2.3cm}}@{}}
            \toprule
            \multirow{2}{*}{Method} & \multirow{2}{*}{Modalities} & \multicolumn{2}{c}{Arousal} & \multicolumn{2}{c}{Valence} & \multicolumn{2}{c}{Four} \\ 
            \cmidrule(l){3-8}
            & &  ACC $\pm$ Std (\%) & F1 $\pm$ Std (\%) &  ACC $\pm$ Std (\%) & F1 $\pm$ Std (\%) & ACC $\pm$ Std (\%) & F1 $\pm$ Std (\%) \\ 
            \midrule
            SimCLR*\cite{chen2020simple}       & EEG            & 92.25 $\pm$ 0.63 & 93.43 $\pm$ 0.56 & 91.39 $\pm$ 0.58 & 92.38 $\pm$ 0.68 & 90.83 $\pm$ 0.91 & 90.82 $\pm$ 0.92 \\ 
            SGMC*\cite{kan2023self}         & EEG            & 95.01 $\pm$ 0.45 & 95.87 $\pm$ 0.42 & 94.56 $\pm$ 0.39 & 94.91 $\pm$ 0.37 & 92.32 $\pm$ 0.47 & 92.35 $\pm$ 0.45 \\ 
            LResCapsule*\cite{fan2024light}  & EEG            & 96.16 $\pm$ 0.38 & 96.98 $\pm$ 0.41 & 95.87 $\pm$ 0.35 & 96.53 $\pm$ 0.32 & 93.96 $\pm$ 0.36 & 93.95 $\pm$ 0.31 \\ 
            CapsNet*\cite{li2022emotion}      & EEG            & 96.70 $\pm$ 1.01 & 97.93 $\pm$ 1.03 & 97.10 $\pm$ 1.23 & 98.04 $\pm$ 1.11 & 96.98 $\pm$ 1.17 & 97.16 $\pm$ 1.09 \\ 
            ODEGC\cite{chen2024eeg}      & EEG            & 97.95 $\pm$ 1.42 & 97.58 ± — & 97.92 $\pm$ 1.74 & 97.48 ± — & — & — \\ 
            DTNET\cite{sun2024emotion}    & EEG       & 98.17 $\pm$ 1.06 & —            & 98.13 $\pm$ 1.12 & —            & —             & —            \\ 
            TACOformer\cite{li2023tacoformer}    & EEG, PPS       & 92.02 $\pm$ 0.73 & —            & 91.59 $\pm$ 0.51 & —            & —             & —            \\ 
            MFST-RNN\cite{li2023emotion}      & EEG, PPS       & 95.89 $\pm$ 1.72 & 96.09 $\pm$ — & 94.99 $\pm$ 1.73 & 95.30 $\pm$ — & —             & —            \\ 
            CAFNet\cite{zhu2023dynamic}        & EEG, Facial    & 94.89 $\pm$ 3.07 & 95.55 $\pm$ 3.30 & 95.25 $\pm$ 3.84 & 94.50 $\pm$ 3.97 & —             & —            \\ 
            
            IANet\cite{li2024incongruity}   & EEG, EOG, EMG, GSR & 97.56 $\pm$ 2.64 & 97.55 $\pm$ 2.64 & 97.42 $\pm$ 1.93 & 97.41 $\pm$ 1.94 & —             & —            \\ 
            MSDSANet\cite{sun2025msdsanet}   & EEG, EOG & 98.19 $\pm$ — & — & 98.07 $\pm$ — & — & —             & —            \\ 
            Ours          & EEG, GSR      & \textbf{98.35 $\pm$ 0.41} & \textbf{98.61 $\pm$ 0.32} & \textbf{98.17 $\pm$ 0.59} & \textbf{98.40 $\pm$ 0.52} & \textbf{97.99 $\pm$ 0.49} & \textbf{97.99 $\pm$ 0.48} \\  
            \bottomrule
        \end{tabular}%
        }
    \end{minipage}
    
\end{table*}

\begin{table*}[!htb]
    \centering
    \begin{minipage}{\textwidth}
        \centering % 表格和描述一起居中
        \caption{Performance comparison with SOAT schemes in terms of accuracy (\%) and F1 score (\%) on the DREAMER dataset.}
        \label{tab:dreamer_sota}
        \setlength{\tabcolsep}{0.3pt} 
        \scalebox{0.85}{%
        \begin{tabular}{@{}cc*{6}{>{\centering\arraybackslash}p{2.3cm}}@{}}
            \toprule
            \multirow{2}{*}{\ \ \ Method} & \multirow{2}{*}{Modalities} & \multicolumn{2}{c}{Arousal} & \multicolumn{2}{c}{Valence} & \multicolumn{2}{c}{Four} \\ 
            \cmidrule(l){3-8}
              & &  ACC $\pm$ Std (\%) & F1 $\pm$ Std (\%) &  ACC $\pm$ Std (\%) & F1 $\pm$ Std (\%) & ACC $\pm$ Std (\%) & F1 $\pm$ Std (\%) \\ 
            \midrule
            \ \ \ SimCLR*\cite{chen2020simple}               & EEG                          & 91.36 ± 0.71 & 92.68 ± 0.64 & 91.11 ± 0.68 & 92.32 ± 0.59 & 90.17 ± 0.37 & 90.15 ± 0.38 \\ 
            \ \ \ SGMC*\cite{kan2023self}                 & EEG                          & 95.35 ± 0.35 & 96.28 ± 0.38 & 94.25 ± 0.37 & 95.12 ± 0.33 & 90.28 ± 0.28 & 91.26 ± 0.35 \\ 
            \ \ \ LResCapsule*\cite{fan2024light}          & EEG                          & 96.39 ± 0.22 & 96.76 ± 0.26 & 94.57 ± 0.29 & 95.13 ± 0.25 & 92.94 ± 0.29 & 92.88 ± 0.31 \\ 
            \ \ \ CapsNet*\cite{li2022emotion}              & EEG                          & 96.02 ± 0.90 & 95.98 ± 0.91 & 95.01 ± 0.88 & 95.48 ± 0.90 & 93.31 ± 0.94 & 93.46 ± 0.98 \\ 
            \ \ \ ODEGC\cite{chen2024eeg}              & EEG                          & 96.01 ± 1.91 & 95.42 ± — & 95.95 ± 1.88 & 94.95 ± — & — & — \\ 
            \ \ \ DTNET\cite{sun2024emotion}             
            & EEG     & 95.17 ± 4.78 & — & 95.62 ± 4.46 & — & — & — \\ 
             \ \ \ DCCA\cite{liu2021comparing}                  & EEG, ECG                     & 89.00 ± 2.80 & —            & 90.60 ± 4.10 & —            & —             & —            \\ 
            \ \ \ TACOformer\cite{li2023tacoformer}            & EEG, ECG                     & 94.03 ± 1.71 & —            & 94.58 ± 4.73 & —            & —             & —            \\ 
           \ \ \ MSDSANet\cite{sun2025msdsanet}            & EEG, ECG                     & 95.54 ± — & —            & 94.83 ± — & —            & —             & —            \\ 
            \ \ \ Ours                  & EEG, ECG                     & \textbf{97.01 ± 0.31} & \textbf{97.55 ± 0.38} & \textbf{96.11 ± 0.32} & \textbf{96.89 ± 0.25} & \textbf{94.70 ± 0.25} & \textbf{94.69 ± 0.25} \\ 
            \bottomrule
        \end{tabular}%
        }
    \end{minipage}
    
\end{table*}

\subsection{Experimental Setup}
All training and experiments were conducted on an NVIDIA TITAN RTX GPU. Due to the limited number of subjects in the dataset, the results in \cite{zhu2023dynamic}, \cite{shen2022contrastive}, \cite{ding2022tsception} show significant data fluctuations when evaluated with cross-subject criteria, with some standard deviations exceeding 10, affecting the reliability. Therefore,we primarily adopt a subject-dependent evaluation using a unified ten-fold cross-validation protocol. In each fold, self-supervised pre-training and fine-tuning were performed exclusively on the corresponding training set, followed by evaluation on the corresponding test set. The best model from each fold was retained, and final results were reported as the mean and standard deviation across all folds. For completeness, we also report results under the cross-subject setting, where the leave-one-subject-out (LOSO) protocol was used. All evaluations strictly follow established practices to ensure reliable and fair comparisons.

With an input signal length of \(t=1s\) for short-term and \(t=5s\) for long-term segments, they are used to train encoders for each time scale during pre-training, with long- and short-term feature extraction performed during fine-tuning. Data augmentation included scaling the samples with factors from [0.7, 0.8] and [1.2, 1.3], and adding Gaussian noise with zero mean to ensure an SNR of 5dB.
% For the hyperparameters, \(\alpha\), \(\beta\), and \(\gamma\) are set to 0.5, 0.5, and 1, respectively.

To optimize the contrastive learning model, we set the number of training epochs to 500, used the Adam optimizer \cite{kingma2014adam}, and applied a cosine annealing learning rate scheduler with triple cyclic warm restarts \cite{loshchilov2016sgdr}. The initial learning rate was set to 0.0001. 
During fine-tuning, the emotion classification task is simpler than feature extraction. To preserve the pre-trained features and avoid overfitting on the small dataset, we set the number of epochs to 15, initial learning rate to 0.001, and batch size to 256, based on empirical observations. Other parameters remained unchanged.

\begin{table*}[h]
    \centering
    \caption{Performance comparison with SOTA schemes in terms of accuracy (\%) and F1 score (\%) on the DEAP and DREAMER datasets under the cross-subject setting.}
    \label{tab:cross-subject}
    \setlength{\tabcolsep}{5pt}
    \scalebox{0.77}{
    \begin{tabular}{@{}c c c c c c c c c@{}}
        \toprule
        \multirow{2}{*}{Method} 
        & \multicolumn{4}{c}{\textbf{DEAP}} 
        & \multicolumn{4}{c}{\textbf{Dreamer}} \\
        \cmidrule(lr){2-5} \cmidrule(lr){6-9}
        & Arousal Acc $\pm$ Std & Arousal F1 $\pm$ Std & Valence Acc $\pm$ Std & Valence F1 $\pm$ Std 
        & Arousal Acc $\pm$ Std & Arousal F1 $\pm$ Std & Valence Acc $\pm$ Std & Valence F1 $\pm$ Std \\
        \midrule
        SimCLR*\cite{chen2020simple}       & 55.26 $\pm$ 10.41 & 60.47 $\pm$ 11.37 & 54.74 $\pm$ 9.63  & 59.19 $\pm$ 10.01 
                       & 60.21 $\pm$ 11.64 & 61.29 $\pm$ 10.23 & 56.91 $\pm$ 11.73 & 60.34 $\pm$ 11.81 \\
        SGMC*\cite{kan2023self}          & 56.35 $\pm$ 11.64 & 61.19 $\pm$ 10.94 & 55.19 $\pm$ 10.44 & 60.81 $\pm$ 9.81  
                       & 61.94 $\pm$ 10.93 & 62.76 $\pm$ 9.98  & 57.02 $\pm$ 12.09 & 61.11 $\pm$ 11.69 \\
        LResCapsule*\cite{fan2024light}   & 55.37 $\pm$ 10.53 & 58.29 $\pm$ 9.42  & 53.36 $\pm$ 12.14 & 54.23 $\pm$ 13.34 
                       & 60.33 $\pm$ 9.89  & 60.22 $\pm$ 12.31 & 55.32 $\pm$ 10.27 & 56.19 $\pm$ 10.68 \\
        CapsNet*\cite{li2022emotion}       & 54.25 $\pm$ 12.64 & 55.26 $\pm$ 11.83 & 52.94 $\pm$ 13.21 & 53.19 $\pm$ 12.18 
                       & 58.24 $\pm$ 11.35 & 58.33 $\pm$ 12.15 & 56.46 $\pm$ 10.62 & 57.51 $\pm$ 11.78 \\
        AD-TCN\cite{he2022adversarial}         & 63.25 $\pm$ 4.64  & —                & 64.33 $\pm$ 7.06  & —                
                       & 63.69 $\pm$ 6.57  & —                & \textbf{66.56 $\pm$ 10.04} & —                \\
        ST-MGRT\cite{xu2025mitigation}        & \textbf{66.21} $\pm$ —     & 60.14 $\pm$ —    & 61.16 $\pm$ —     & 57.46 $\pm$ —     
                       & \textbf{66.22} $\pm$ —     & 60.35 $\pm$ —    & 61.16 $\pm$ —     & 57.46 $\pm$ —     \\
        EmotionMIL\cite{xiao2025emotionmil}     & 56.88 $\pm$ 12.43 & 52.97 $\pm$ 16.03 & 57.66 $\pm$ 14.74 & 48.57 $\pm$ 17.90 
                       & 60.63 $\pm$ 10.97 & 47.34 $\pm$ 11.88 & 57.49 $\pm$ 15.32 & 50.94 $\pm$ 15.62 \\
        Ours  & 65.46 $\pm$ 8.15 & \textbf{64.33 $\pm$ 7.01} & \textbf{64.79 $\pm$ 7.24} & \textbf{64.10 $\pm$ 6.35} 
                       &64.88 $\pm$ 8.06 & \textbf{65.89 $\pm$ 7.37} & 63.24 $\pm$ 8.56 & \textbf{62.57 $\pm$ 8.12} \\
        \bottomrule
    \end{tabular}
    }
\end{table*}

% \begin{figure}[h] % figure for single column
%     \centering
%     \includegraphics[width=\columnwidth]{./finetuning.png} % replace example-image with your image file name
%     \caption{The illustration of the fine-tuning strategy without long-term and short-term feature fusion.}
%     \label{fig:example}
% \end{figure}

\begin{figure*}[!htb]
    \centering
    \includegraphics[width=0.75\textwidth, height=0.42\textwidth]{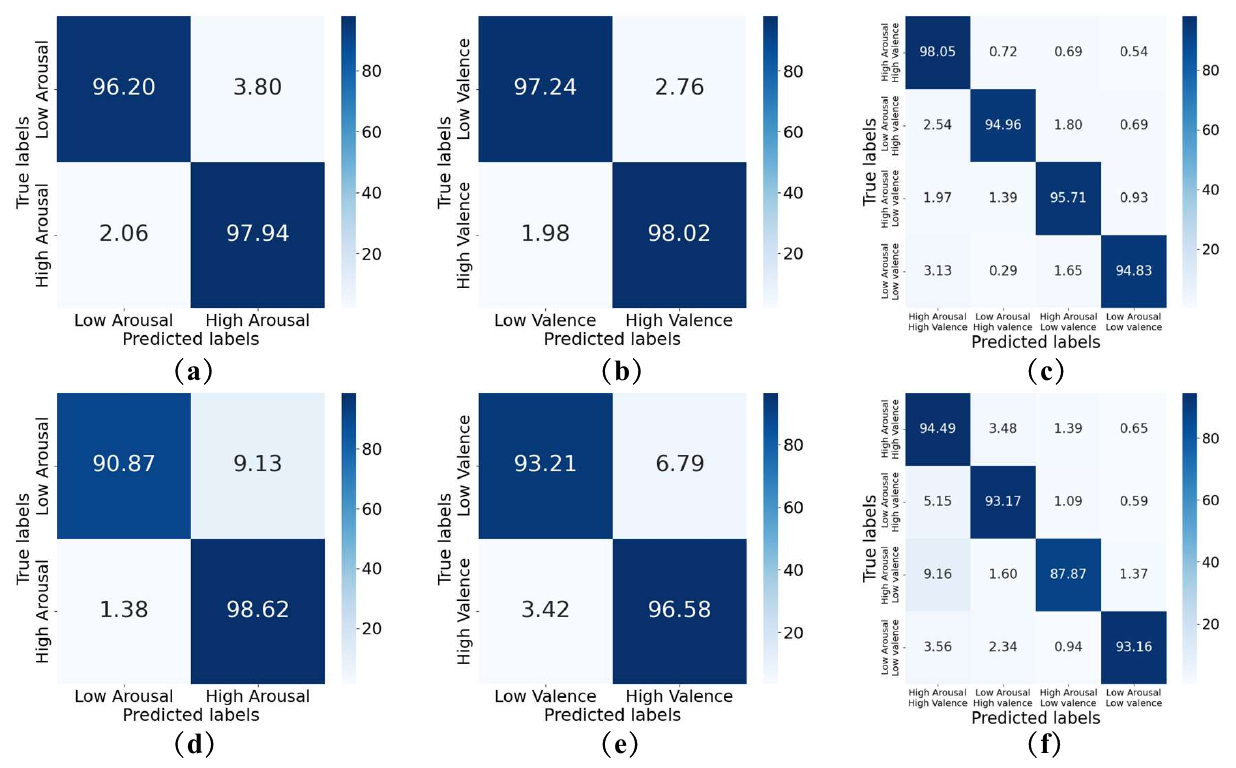}
    \caption{Confusion matrices in the DEAP and DREAMER datasets. (a) DEAP-Arousal. (b) DEAP-Valence. (c) DEAP-Four. (d) DREAMER-Arousal. (e) DREAMER-Valence. (f) DREAMER-Four.}
    \label{fig:cm}
\end{figure*}

\begin{figure*}[!htb]
    \centering
    \includegraphics[width=0.83\textwidth, height=0.26\textwidth]{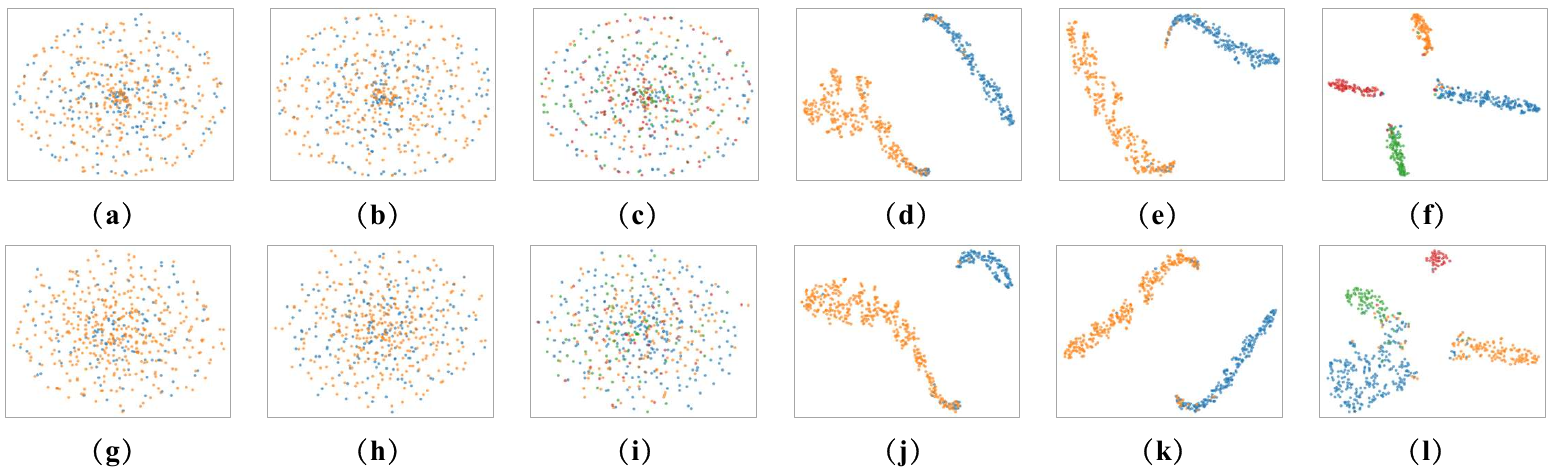}
    \caption{T-SNE visualization in the DEAP and DREAMER datasets. The original feature distributions of arousal, valence, and the four-class for the DEAP and DREAMER datasets are represented by (a)-(c) and (g)-(i), respectively. The feature distributions of our method across various dimensional spaces for both datasets are shown in (d)-(f) and (g)-(l). In the two-color image, \(\textcolor[rgb]{0.39, 0.58, 0.93}{\huge \bullet}\) \(\textcolor[rgb]{1.00, 0.65, 0.31}{\huge \bullet}\) represent the low and high arousal (or valence), respectively. In the four-color image,  \(\textcolor[rgb]{0.39, 0.58, 0.93}{\huge \bullet}\) \(\textcolor[rgb]{1.00, 0.65, 0.31}{\huge \bullet}\) \(\textcolor[rgb]{0.79, 0.26, 0.19}{\huge \bullet}\) \(\textcolor[rgb]{0.13, 0.54, 0.13}{\huge \bullet}\) represent high arousal/high valence, low arousal/high valence, high arousal/low valence and low arousal/low valence, respectively.}
    \label{fig:t_sne}
\end{figure*}

\begin{figure}[htbp]
    \centering
    \includegraphics[width=\columnwidth, height=0.13\textheight]{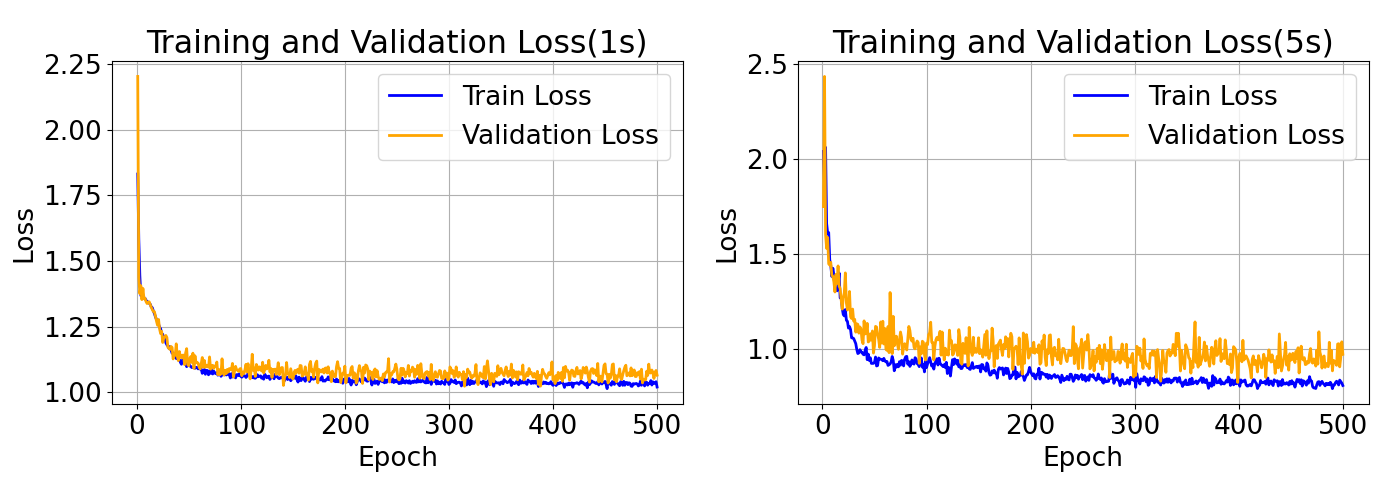}
    \caption{Training and validation loss curves over epochs for both 1-second and 5-second windows during the pre-training phase.}
    \label{fig:loss_curve}
\end{figure}

\begin{table*}[h]
    \centering
    \begin{minipage}{\textwidth}
        \centering
        \caption{Ablation experiments on the DEAP dataset. TCL represents intra-modal temporal contrastive learning, DA represents data augmentation, CM-CL represents cross-modal contrastive learning.}
        \label{tab:abl_structure}
        \setlength{\tabcolsep}{6pt} 
        \scalebox{0.88}{%
        \begin{tabular}{@{}cccccccccc@{}}
            \toprule
            \multirow{2}{*}{\ \ \ Method} & \multirow{2}{*}{TCL} & \multirow{2}{*}{DA} & \multirow{2}{*}{CM-CL} & \multicolumn{2}{c}{Arousal} & \multicolumn{2}{c}{Valence} & \multicolumn{2}{c}{Four} \\ 
            \cmidrule(lr){5-10}
            & & & & ACC $\pm$ Std (\%) & F1 $\pm$ Std (\%) & ACC $\pm$ Std (\%) & F1 $\pm$ Std (\%) & ACC $\pm$ Std (\%) & F1 $\pm$ Std (\%) \\ 
            \midrule
            \ \ \ Ours-1 &  &  &  & 95.96 $\pm$ 0.71 & 97.01 $\pm$ 0.73 & 95.91 $\pm$ 0.60 & 96.90 $\pm$ 0.78 & 95.32 $\pm$ 0.51 & 95.31 $\pm$ 0.50 \\
            \ \ \ Ours-2 & $\surd$ &  &  & 96.99 $\pm$ 0.51 & 97.23 $\pm$ 0.49 & 97.03 $\pm$ 0.56 & 97.58 $\pm$ 0.51 & 96.15 $\pm$ 0.63 & 96.10 $\pm$ 0.75 \\
            \ \ \ Ours-3 &  & $\surd$ &  & 97.01 $\pm$ 0.36 & 97.56 $\pm$ 0.39 & 97.06 $\pm$ 0.41 & 97.41 $\pm$ 0.46 & 96.22 $\pm$ 0.38 & 96.15 $\pm$ 0.34 \\
            \ \ \ Ours-4 &  &  & $\surd$ & 96.68 $\pm$ 0.65 & 97.15 $\pm$ 0.74 & 96.65 $\pm$ 0.60 & 97.10 $\pm$ 0.58 & 95.86 $\pm$ 0.51 & 95.80 $\pm$ 0.54 \\
            \ \ \ Ours-5 & $\surd$ & $\surd$ &  & 97.35 $\pm$ 0.46 & 97.63 $\pm$ 0.48 & 97.16 $\pm$ 0.41 & 97.56 $\pm$ 0.43 & 96.92 $\pm$ 0.43 & 96.93 $\pm$ 0.44 \\
            \ \ \ Ours-6 & $\surd$ &  & $\surd$ & 97.54 $\pm$ 0.51 & 97.74 $\pm$ 0.49 & 97.63 $\pm$ 0.57 & 97.69 $\pm$ 0.58 & 96.64 $\pm$ 0.71 & 96.63 $\pm$ 0.70 \\
            \ \ \ Ours-7 &  & $\surd$ & $\surd$ & 97.39 $\pm$ 0.48 & 97.84 $\pm$ 0.45 & 97.31 $\pm$ 0.37 & 97.88 $\pm$ 0.39 & 96.96 $\pm$ 0.39 & 96.97 $\pm$ 0.34 \\
            \ \ \ Ours & $\surd$ & $\surd$ & $\surd$ & \textbf{97.92 $\pm$ 0.51} & \textbf{98.25 $\pm$ 0.38} & \textbf{98.01 $\pm$ 0.42} & \textbf{98.22 $\pm$ 0.38} & \textbf{97.65 $\pm$ 0.32} & \textbf{97.66 $\pm$ 0.31} \\
            \bottomrule
        \end{tabular}%
        }
    \end{minipage}
\end{table*}

\begin{table}[h!]
    \centering
    \caption{Comparison of different data augmentation methods in DEAP dataset. SN:Scaling and Noise, CP:Channel Permutation, TF:Temporal Flipping, 
    All:All combined augmentations.}
    \label{tab:compare_DA_methods}
    \setlength{\tabcolsep}{5pt}
    \scalebox{0.79}{%
    \begin{tabular}{c c c c c}
        \toprule
        \multirow{2}{*}{\textbf{Method}} & \multicolumn{2}{c}{\textbf{Arousal}} & \multicolumn{2}{c}{\textbf{Valence}} \\ 
        \cmidrule(lr){2-3} \cmidrule(lr){4-5}
        & ACC $\pm$ Std (\%) & F1 $\pm$ Std (\%) & ACC $\pm$ Std (\%) & F1 $\pm$ Std (\%) \\ 
        \midrule
        Baseline         & 97.54 $\pm$ 0.51 & 97.74 $\pm$ 0.49 & 97.63 $\pm$ 0.57 & 97.69 $\pm$ 0.58 \\
        SN & 97.92 $\pm$ 0.51 & 98.25 $\pm$ 0.38 & 98.01 $\pm$ 0.42 & 98.22 $\pm$ 0.38 \\
        SN+CP & 97.96 $\pm$ 0.63 & \textbf{98.29 $\pm$ 0.47} & \textbf{98.08 $\pm$ 0.41} & 98.25 $\pm$ 0.49 \\
        SN+TF       & 97.88 $\pm$ 0.58 & 98.12 $\pm$ 0.39 & 98.06 $\pm$ 0.61 & 98.19 $\pm$ 0.53 \\
        All        & \textbf{98.01 $\pm$ 0.61} & 98.26 $\pm$ 0.55 & 98.05 $\pm$ 0.48 & \textbf{98.29 $\pm$ 0.44} \\
        \bottomrule
    \end{tabular}%
    }
\end{table}

\subsection{Subject-Dependent Performance Comparison with SOTA}
To demonstrate the superiority of the proposed method, we conducted extensive performance comparisons on the DEAP and DREAMER datasets. First, we selected several EEG-only models, results for these "*" models were obtained via ten-fold cross-validation using published source code and consistent experimental criteria. Additionally, we compared several multimodal fusion models. 

As shown in Tables \ref{tab:deap_sota} and \ref{tab:dreamer_sota}, our proposed method outperforms all others. Specifically, we obtain average accuracies of 98.35\%, 98.17\%, and 97.99\% on DEAP, and 97.01\%, 96.11\%, and 94.70\% on DREAMER, with corresponding F1 scores closely aligned.
Compared to models using EEG alone, our method achieves a recognition accuracy increase of 0.18\%  (arousal), 0.04\% (valence), and 1.31\%  (four-class) on the DEAP dataset. In the DREAMER dataset, the accuracy increases by 0.62\%, 0.16\%, and 1.39\%, demonstrating the benefit of fusing EEG with peripheral physiological signals. These gains highlight the complementary nature of the two modalities in capturing richer emotional information. 

Further, against existing multimodal methods, our approach achieves additional improvements—up to 0.16\% on DEAP and 1.47\% on DREAMER in arousal,  0.10\% and 1.28\% in valence—indicating better exploitation of cross-modal information and more effective suppression of redundancy. Similar trends are observed in the F1 scores, confirming consistent performance gains. Moreover, both the accuracy and F1 scores have a standard deviation (STD) not exceeding 0.6. This low variability is primarily due to the use of subject-dependent experiments, where training and testing are performed on data from the same individual. Such settings typically yield more stable and consistent results compared to cross-subject scenarios. In summary, our approach not only enhances performance but also improves the reliability and stability of emotion recognition using multimodal physiological signals.

\subsection{Cross-Subject Performance Comparison with SOTA}
To evaluate the generalization ability of our method, we conducted cross-subject experiments on the DEAP and DREAMER datasets using the Leave-One-Subject-Out (LOSO) protocol. As shown in Table~\ref{tab:cross-subject}.   Specifically, it reaches 65.46\% / 64.79\% (Acc) and 64.33\% / 64.10\% (F1) on DEAP, and 64.88\% / 63.24\% (Acc) and 65.89\% / 62.57\% (F1) on DREAMER. Although the corresponding accuracy does not reach the absolute highest, the results remain highly competitive, and the consistently strong F1 scores reflect better balance between precision and recall, indicating more robust generalization across subjects. This suggests that the model is less affected by individual variability and can effectively learn subject-invariant representations.

\subsection{Qualitative Analysis}
In this section, we visually analyze the challenges of emotion recognition. Fig. \ref{fig:cm} shows the confusion matrices of our model on the DEAP and DREAMER datasets, demonstrating its ability to distinguish emotional states. Higher accuracy is observed in both arousal and valence dimensions for high values, typically linked to discrete positive emotions like happiness and excitement. These emotions correspond to distinct physiological signals—such as increased heart rate, skin conductance, and brain activity—making them easier to recognize.
When combining arousal and valence into four categories, the best performance occurs in the high valence/high arousal group for both datasets, consistent with previous binary classification results.

To demonstrate the model’s classification ability, we applied t-SNE for dimensionality reduction to compare raw features and features after training (see Fig. \ref{fig:t_sne}).
The trained features show significantly improved separation between emotion categories, with clear clustering across all dimensions. This indicates enhanced feature extraction and classification performance, allowing the model to better capture and distinguish latent emotional patterns.
Additionally, the trained features exhibit a continuous band-like distribution, reflecting the smoothness of emotion labels—that is, their gradual changes over time. Although labels are binarized for classification, this pattern shows the model captures underlying temporal dynamics beyond discrete categories.

To further demonstrate the convergence and stability of the model during pre-training, we plot the training and validation loss curves for both 1-second and 5-second windows in Fig.~\ref{fig:loss_curve}. As shown, the training and validation losses decrease steadily and plateau without signs of divergence, indicating effective learning and generalization. This supports the robustness of the proposed model across varying temporal resolutions.

\section{Ablation Experiments}
This section presents a series of ablation experiments. Notably, to control for confounding factors, all experiments during fine-tuning utilized only long-term features, except for the ablation experiment of the long short term strategy.

% \subsection{Effectiveness of Key Modules}
\subsection{Analysis of Data Augmentation and Contrastive Learning}
To validate the effectiveness of each module, we conducted multiple ablation experiments by sequentially adding each module (see Table \ref{tab:abl_structure}). 
Three components were selected for ablation studies: intra-modal temporal contrastive learning (TCL), data augmentation (DA), and cross-modal contrastive learning (CM-CL). 
By applying intra-modal temporal contrastive learning, the encoders can extract physiological features consistent with the stimuli, facilitating the distinction of emotional states and achieving physiological synchronization across subjects. Data augmentation increases training samples and positive pairs, enhancing contrastive learning and improving model generalization. Cross-modal contrastive learning strengthens interaction and alignment between modalities, mapping signals into a shared semantic space for physiological synchronization. Results show that recognition performance steadily improves as each module is added, confirming the effectiveness of our approach.

Additionally, we explored other common data augmentation techniques beyond scaling and noise injection, including channel permutation (CP), and signal temporal flipping (TF). The ablation results are summarized in \ref{tab:compare_DA_methods}. Compared with the baseline method (Baseline: no augmentation), each augmentation method contributes to improved performance in both arousal and valence dimensions. Notably,  method All achieves competitive but not the best results, indicating that more augmentations do not necessarily lead to better performance.

\begin{figure*}[!htb]
    \centering
    \includegraphics[width=0.68\textwidth, height=0.23\textwidth]{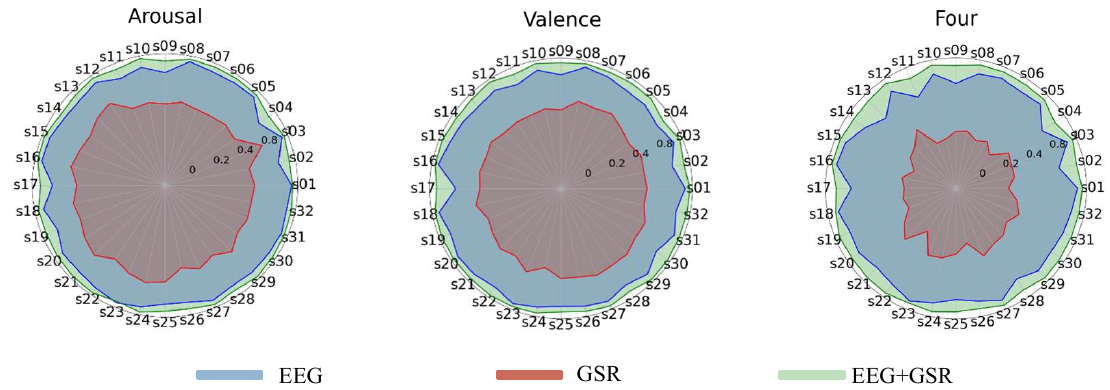}
    \caption{The recognition accuracy of arousal, valence, and four dimensions of each subject in DEAP dataset by using EEG, GSR, and modality fusion.}
    \label{fig:sub}
\end{figure*}

% \subsection{Ablation of Different Modalities}
\subsection{Ablation of EEG and PPS} 
To validate the complementary role of PPS to EEG, we conducted a modality ablation experiment on the DEAP dataset. We tested four modalities—EEG, Electromyography (EMG), Electrooculography (EOG), and GSR—both individually and in combination, as summarized in Table see Figure \ref{tab:abl_modality}. The results show that incorporating peripheral physiological signals improved recognition accuracy by 2.91\% in arousal and 3.03\% in valence, while F1 scores increased by 1.47\% in arousal and 1.66\% in valence.
Specifically, the combination of EEG and GSR yielded the best results across most dimensions. 
The SOTA comparison experiment results are based on the fusion of EEG and GSR.
Among the single modalities, EEG consistently achieved the highest accuracy and F1 scores across the Arousal, Valence, and four-category tasks, further emphasizing its critical role in emotion recognition. The relatively limited contribution from EMG and EOG may be due to their weaker correlation with emotional states—EMG is more susceptible to motion artifacts\cite{golland2018affect}, while EOG mainly reflects eye movements, which not be strongly linked to arousal or valence\cite{simola2015affective}.

\begin{table}[h!]
    \centering
    \begin{minipage}{\columnwidth} % 设置为一栏的宽度
        \caption{Comparison of the performance of single modality and pairwise different modality combinations in DEAP dataset.}
        \label{tab:abl_modality}
        \scalebox{0.78}{%
        \begin{tabular}{c c c c c}
            \toprule
            \multirow{2}{*}{\textbf{Modality}} & \multicolumn{2}{c}{\textbf{Arousal}} & \multicolumn{2}{c}{\textbf{Valence}} \\ 
            \cmidrule(lr){2-3} \cmidrule(lr){4-5}
            & ACC $\pm$ Std (\%) & F1 $\pm$ Std (\%) & ACC $\pm$ Std (\%) & F1 $\pm$ Std (\%) \\ 
            \midrule
            EEG      & 95.01 $\pm$ 0.47 & 96.78 $\pm$ 0.45 & 94.98 $\pm$ 0.28 & 96.56 $\pm$ 0.44 \\ 
            EMG      & 59.93 $\pm$ 1.20 & 62.30 $\pm$ 1.98 & 60.15 $\pm$ 1.45 & 63.91 $\pm$ 1.68 \\ 
            EOG      & 64.70 $\pm$ 1.83 & 68.56 $\pm$ 1.89 & 62.05 $\pm$ 1.95 & 67.86 $\pm$ 2.11 \\ 
            GSR      & 63.19 $\pm$ 2.29 & 66.76 $\pm$ 2.86 & 63.09 $\pm$ 2.35 & 65.53 $\pm$ 2.18 \\ 
            \midrule
            EEG + GSR & \textbf{97.92 $\pm$ 0.51} & \textbf{98.25 $\pm$ 0.38} & \textbf{98.01 $\pm$ 0.42} & \textbf{98.22 $\pm$ 0.38} \\ 
            EEG + EMG & 97.73 $\pm$ 0.54 & 98.08 $\pm$ 0.48 & 97.77 $\pm$ 0.59 & 98.03 $\pm$ 0.55 \\ 
            EEG + EOG & 97.85 $\pm$ 0.46 & 98.19 $\pm$ 0.39 & 97.55 $\pm$ 0.59 & 97.82 $\pm$ 0.60 \\ 
            \bottomrule
        \end{tabular}%
        }
    \end{minipage}
\end{table}

We further conducted experiments using the EEG and GSR data from each subject separately to validate the role of modality fusion. As shown in Fig. \ref{fig:sub}, the fusion of EEG and GSR (green line) consistently outperformed the single modalities (EEG in blue and GSR in red) across all 32 subjects in the arousal, valence, and four-class classification tasks. This demonstrates the complementary nature of EEG and GSR, highlighting the strength of modality fusion in providing a richer and more robust representation of emotional states.

\begin{figure*}[!htb]
    \centering
    \includegraphics[width=0.77\textwidth, height=0.24\textwidth ]{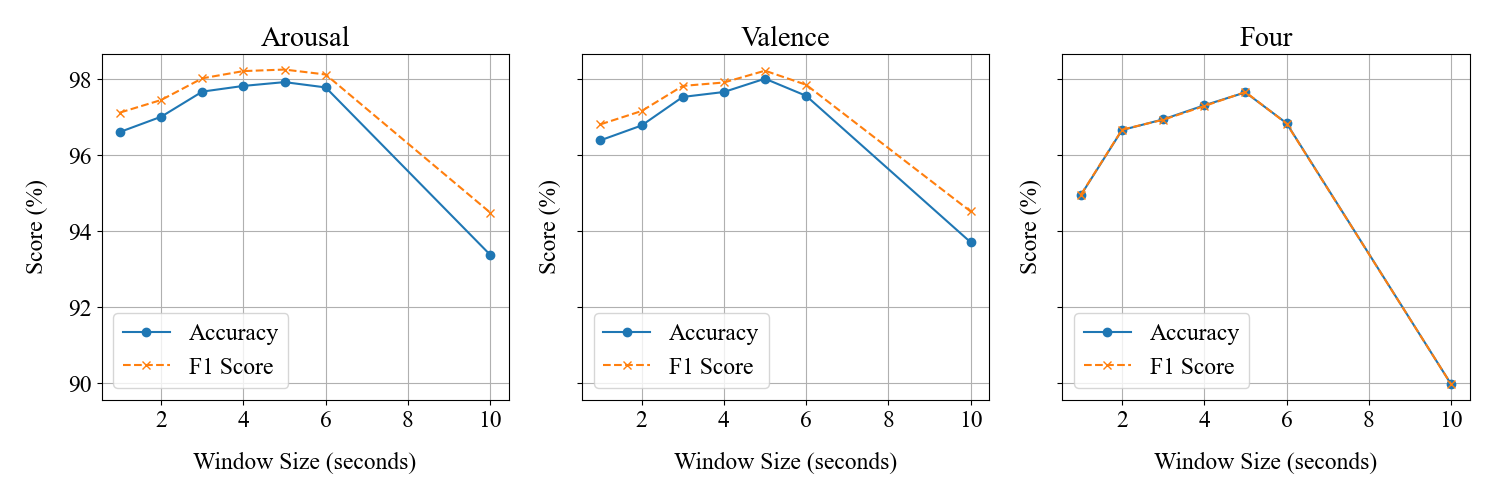}
    \caption{Illustration of model performance with different time windows on the DEAP dataset. The best performance is achieved with a 5-second window.}
    \label{fig:window}
\end{figure*}

% \subsection{Ablation of Long Short Term Strategy}
% \subsection{Ablation of Long-Term and Short-Term Feature Fusion}
\subsection{Ablation of Long-Term Signal Decomposition for Short-Term Fusion}
In the final fine-tuning strategy, we combined long-term and short-term temporal features. 
Here, ``Long" refers to using only long-term features,``Long+Short" uses a fusion of both long- and short-term features, and ``Long*" adds two self-attention layers on top of the "Long" strategy. 
Specifically(see Table \ref{tab:abl_short_long}), compared to using long-term features alone, further fusion with short-term features improves recognition accuracy by 0.43\% in arousal and 0.16\% in valence, and F1 scores increased by 0.43\% in arousal and 0.18\% in valence.
The performance improvement highlights the value of multi-scale feature integration in emotion recognition. In contrast, adding parameters with the ``Long*" strategy alone does not improve performance, indicating that relying solely on long-term features, even with more parameters, limits the model's capacity to capture dynamic emotional changes.

\begin{table}[h!]
    \centering
    \begin{minipage}{\columnwidth} % 设置为一栏宽度
        \caption{Comparison of classification results for long-time features and fusion of long-short time features.}
        \label{tab:abl_short_long}
        \scalebox{0.72}{%
        \begin{tabular}{c c c c c c}
            \toprule
            \multirow{2}{*}{\textbf{Method}} & \multirow{2}{*}{\textbf{Params}} & \multicolumn{2}{c}{\textbf{Arousal}} & \multicolumn{2}{c}{\textbf{Valence}} \\ 
            \cmidrule(lr){3-4} \cmidrule(lr){5-6}
            & & ACC $\pm$ Std (\%) & F1 $\pm$ Std (\%) & ACC $\pm$ Std (\%) & F1 $\pm$ Std (\%) \\ 
            \midrule
            Long         & 116M & 97.92 $\pm$ 0.51 & 98.25 $\pm$ 0.38 & 98.01 $\pm$ 0.42 & 98.22 $\pm$ 0.38 \\ 
            Long*        & 141M & 97.35 $\pm$ 0.45 & 97.86 $\pm$ 0.54 & 97.21 $\pm$ 0.47 & 97.74 $\pm$ 0.46 \\ 
            Long + Short & 142M & \textbf{98.35 $\pm$ 0.41} & \textbf{98.61 $\pm$ 0.32} & \textbf{98.17 $\pm$ 0.59} & \textbf{98.40 $\pm$ 0.52} \\ 
            \bottomrule
        \end{tabular}%
        }
    \end{minipage}
\end{table}

% \subsection{Ablation of Various Time Resolutions(Windows)}
\subsection{Impact of Time Resolutions (Windows)}
In this section, we systematically examine the effect of time window length on model performance in emotion recognition using physiological signals (see Fig. \ref{fig:window}). Signal segments ranging from 1 to 10 seconds (excluding 7s–9s) were evaluated. Results show a clear trend: performance improves notably from 1 to 5 seconds, with both accuracy and F1 scores increasing as longer windows capture richer physiological features. The best performance is achieved at 5 seconds, which offers an optimal trade-off between accuracy and computational efficiency. However, when the window length exceeds 6 seconds, performance begins to decline. This may be due to emotional drift\cite{chanel2009short}, where the subject’s emotional state changes within the segment, introducing label noise. Additionally, longer segments may contain redundant or diluted information, making it harder for the model to capture key discriminative features. Overall, a 5-second window strikes the optimal trade-off between performance and efficiency.

% \textcolor{red}{we performed a misalignment experiment by intentionally shifting PPS signals relative to EEG. The results in \ref{tab:mis_align} showed a clear degradation in performance under these conditions, confirming that proper synchronization across modalities is essential for effective feature learning. This result supports the motivation behind our physiological synchronization strategy, suggesting that temporal alignment plays an important role in cross-modal learning.}

% \begin{table}[h!]
%     \centering % 表格居中
%     \begin{minipage}{\columnwidth} % 设置为一栏宽度
%         \caption{\textcolor{red}{Comparison of synchronized and misaligned inputs.}}
%         \label{tab:mis_align}
%         \scalebox{0.87}{%
%         \begin{tabular}{c c c c c}
%             \toprule
%             \multirow{2}{*}{\textbf{Method}} & \multicolumn{2}{c}{\textbf{Arousal}} & \multicolumn{2}{c}{\textbf{Valence}} \\ 
%             \cmidrule(lr){2-3} \cmidrule(lr){4-5}
%             & ACC $\pm$ Std (\%) & F1 $\pm$ Std (\%) & ACC $\pm$ Std (\%) & F1 $\pm$ Std (\%) \\ 
%             \midrule
%             w/ Sync & 97.92 $\pm$ 0.51 & 98.25 $\pm$ 0.38 & 98.01 $\pm$ 0.42 & 98.22 $\pm$ 0.38 \\ 
%             w/o Sync & 96.12 $\pm$ 1.02 & 96.33 $\pm$ 0.94 & 96.03 $\pm$ 0.91 & 96.21 $\pm$ 0.95 \\ 
%             \bottomrule
%         \end{tabular}%
%         }
%     \end{minipage}
% \end{table}

% \subsection{Comparison of Pair Selection Strategies}
\subsection{Comparison of Physiological Signal Pairing Strategies}
To validate the effectiveness of our positive-negative pair selection strategy, we compared it with CLISA\cite{shen2022contrastive}, the best-performing single-modality method using only EEG.
In CLISA, 5-second segments are randomly selected from each trial and paired across subjects as positive pairs. Our method pre-divides each trial into chronological 5-second segments, randomly selects K segments, and generates negative pairs within the same trial. This increases the challenge of contrastive learning through more refined sample pairing. 
As shown in Table \ref{tab:pair} (Our* means that our method does not use data augmentation), our construction method better helps the model distinguish between different emotional categories. Specifically, accuracy and F1 scores improved by 0.34\% and 1.54\% in arousal, 0.38\% and 1.35\% in valence, while corresponding standard deviations decreased by \(0.06\sim0.41\) and \(0.21\sim0.37\), respectively. Our intra-trial negative pairing surpasses CLISA by learning fine-grained within-subject emotional differences, boosting robustness and discriminability.

\begin{table}[h!]
    \centering % 表格居中
    \begin{minipage}{\columnwidth} % 设置为一栏宽度
        \caption{Performance of pair selection strategies}
        \label{tab:pair}
        \scalebox{0.87}{%
        \begin{tabular}{c c c c c}
            \toprule
            \multirow{2}{*}{\textbf{Method}} & \multicolumn{2}{c}{\textbf{Arousal}} & \multicolumn{2}{c}{\textbf{Valence}} \\ 
            \cmidrule(lr){2-3} \cmidrule(lr){4-5}
            & ACC $\pm$ Std (\%) & F1 $\pm$ Std (\%) & ACC $\pm$ Std (\%) & F1 $\pm$ Std (\%) \\ 
            \midrule
            CLSIA\cite{shen2022contrastive} & 94.67 $\pm$ 0.53 & 95.26 $\pm$ 0.67 & 94.60 $\pm$ 0.72 & 95.21 $\pm$ 0.81 \\ 
            Ours$^*$ & 94.81 $\pm$ 0.51 & 95.35 $\pm$ 0.59 & 94.55 $\pm$ 0.39 & 95.85 $\pm$ 0.62 \\ 
            Ours & \textbf{95.01 $\pm$ 0.47} & \textbf{96.78 $\pm$ 0.45} & \textbf{94.98 $\pm$ 0.41} & \textbf{96.56 $\pm$ 0.44} \\ 
            \bottomrule
        \end{tabular}%
        }
    \end{minipage}
\end{table}

% \subsection{Ablation of Fusion Methods}
\subsection{Ablation of Modal Fusion Strategies}
\begin{figure}[h] 
    \centering
    \includegraphics[width=\columnwidth, height=0.21\textheight]{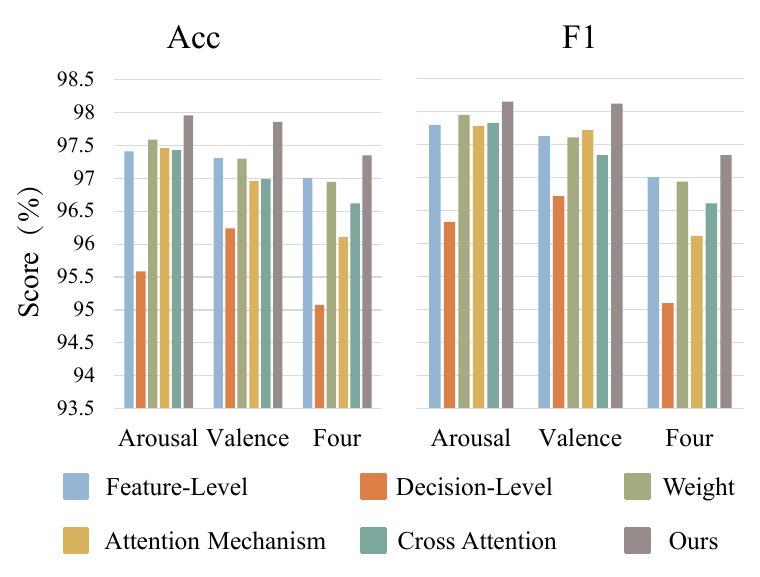} % Set both width and height
    \caption{Comparison of different modal fusion methods (Including feature-level fusion, decision-level fusion, adaptive weighting, attention mechanisms, and cross-attention).}
    \label{fig:fusion}
\end{figure}

In downstream tasks involving the fusion of different modalities, we compared several common fusion methods, including feature-level fusion\cite{2009Feature}, decision-level fusion\cite{zhang2009decision}, adaptive weighting\cite{xu2015adaptive}, attention mechanisms\cite{dai2021attentional}, and cross-attention\cite{hou2019cross} (see Fig. \ref{fig:fusion}). The decision-level fusion combines the final outputs of the classification head, while other methods perform fusion on the features extracted by the encoder. The results show that our fusion method outperforms others in terms of performance. By assigning a weight to each modality based on its maximum class probability before fusion, our method ensures an appropriate contribution from each modality, enabling the model to prioritize the most informative signals while still considering all modalities.

\section{LIMITATION}
In this study, we primarily adopted a subject-dependent evaluation protocol due to large performance fluctuations observed under the cross-subject setting, with standard deviations exceeding 10 in some cases. This issue is largely attributed to the limited number of subjects and data in the datasets. Although we included cross-subject experimental results, a detailed analysis was not performed in this work. In future studies, we plan to explore this setting more thoroughly as an important research direction. In addition, our current model only supports fusion of two modalities. In future work, we will explore fusion of multiple modalities to better integrate complementary information from other physiological signals.

\section{CONCLUSION}
In this paper, inspired by physiological synchronization, we propose a contrastive learning framework for emotion recognition, achieving competitive results on the DEAP and DREAMER datasets. By integrating intra-modal temporal contrastive learning (TCL) and cross-modal contrastive learning (CM-CL), we make preliminary attempts at achieving physiological synchronization across both subjects and modalities. 
Additionally, we found that a 5-second time window performed optimally when using a single time resolution feature.
However, as emotions are characterized by both instantaneous and subtle changes, further extraction and fusion of long-term and short-term features from different time resolutions have led to even better results.
Experiments show that, although EEG signals generally perform well in emotion recognition, a reasonable and effective combination with peripheral physiological signals can lead to even better results.
Physiological signals are typically multi-channel, with spatial relationships between channels that our model has not yet fully exploited. Therefore, in addition to improving the cross-subject experiments and exploring fusion of multiple modalities, future research will focus on better exploring and utilizing these spatial relationships to further enhance model performance.
% \newpage

\bibliographystyle{unsrt}
\bibliography{myref}

\begin{thebibliography}{10}

\bibitem{zhang2023discriminative}
Xiaowei Zhang, Zhongyi Zhou, Qiqi Zhao, Kechen Hou, Xiangyu Wei, Sipo Zhang,
  Yikun Yang, and Yanmeng Cui.
\newblock Discriminative joint knowledge transfer with online updating
  mechanism for eeg-based emotion recognition.
\newblock {\em IEEE Transactions on Computational Social Systems},
  11(2):2918--2929, 2023.

\bibitem{jafari2023emotion}
Mahboobeh Jafari, Afshin Shoeibi, Marjane Khodatars, Sara Bagherzadeh, Ahmad
  Shalbaf, David~L{\'o}pez Garc{\'\i}a, Juan~M Gorriz, and U~Rajendra Acharya.
\newblock Emotion recognition in eeg signals using deep learning methods: A
  review.
\newblock {\em Computers in Biology and Medicine}, 165:107450, 2023.

\bibitem{alarcao2017emotions}
Soraia~M Alarcao and Manuel~J Fonseca.
\newblock Emotions recognition using eeg signals: A survey.
\newblock {\em IEEE transactions on affective computing}, 10(3):374--393, 2017.

\bibitem{mauss2009measures}
Iris~B Mauss and Michael~D Robinson.
\newblock Measures of emotion: A review.
\newblock {\em Cognition and emotion}, 23(2):209--237, 2009.

\bibitem{li2022sstd}
Rui Li, Chao Ren, Chen Li, Nan Zhao, Dawei Lu, and Xiaowei Zhang.
\newblock Sstd: a novel spatio-temporal demographic network for eeg-based
  emotion recognition.
\newblock {\em IEEE Transactions on Computational Social Systems},
  10(1):376--387, 2022.

\bibitem{li2024gusa}
Xiaojun Li, CL~Philip Chen, Bianna Chen, and Tong Zhang.
\newblock Gusa: Graph-based unsupervised subdomain adaptation for cross-subject
  eeg emotion recognition.
\newblock {\em IEEE Transactions on Affective Computing}, 2024.

\bibitem{ye2024adaptive}
Mengqing Ye, CL~Philip Chen, Wenming Zheng, and Tong Zhang.
\newblock Adaptive dual-space network with multigraph fusion for eeg-based
  emotion recognition.
\newblock {\em IEEE Transactions on Computational Social Systems}, 2024.

\bibitem{gu2022multi}
Xiaoqing Gu, Weiwei Cai, Ming Gao, Yizhang Jiang, Xin Ning, and Pengjiang Qian.
\newblock Multi-source domain transfer discriminative dictionary learning
  modeling for electroencephalogram-based emotion recognition.
\newblock {\em IEEE Transactions on Computational Social Systems},
  9(6):1604--1612, 2022.

\bibitem{tang2024hierarchical}
Jiehao Tang, Zhuang Ma, Kaiyu Gan, Jianhua Zhang, and Zhong Yin.
\newblock Hierarchical multimodal-fusion of physiological signals for emotion
  recognition with scenario adaption and contrastive alignment.
\newblock {\em Information Fusion}, 103:102129, 2024.

\bibitem{shen2024tensor}
Jian Shen, Kexin Zhu, Huakang Liu, Jinwen Wu, Kang Wang, and Qunxi Dong.
\newblock Tensor correlation fusion for multimodal physiological signal emotion
  recognition.
\newblock {\em IEEE Transactions on Computational Social Systems}, 2024.

\bibitem{zhu2023dynamic}
Qi~Zhu, Chuhang Zheng, Zheng Zhang, Wei Shao, and Daoqiang Zhang.
\newblock Dynamic confidence-aware multi-modal emotion recognition.
\newblock {\em IEEE Transactions on Affective Computing}, 2023.

\bibitem{fu2024cross}
Baole Fu, Wenhao Chu, Chunrui Gu, and Yinhua Liu.
\newblock Cross-modal guiding neural network for multimodal emotion recognition
  from eeg and eye movement signals.
\newblock {\em IEEE Journal of Biomedical and Health Informatics}, 2024.

\bibitem{arthanarisamy2022subject}
Manju~Priya Arthanarisamy~Ramaswamy and Suja Palaniswamy.
\newblock Subject independent emotion recognition using eeg and physiological
  signals--a comparative study.
\newblock {\em Applied Computing and Informatics}, 2022.

\bibitem{hasson2004intersubject}
Uri Hasson, Yuval Nir, Ifat Levy, Galit Fuhrmann, and Rafael Malach.
\newblock Intersubject synchronization of cortical activity during natural
  vision.
\newblock {\em science}, 303(5664):1634--1640, 2004.

\bibitem{stuldreher2020physiological}
Ivo~V Stuldreher, Nattapong Thammasan, Jan~BF van Erp, and Anne-Marie Brouwer.
\newblock Physiological synchrony in eeg, electrodermal activity and heart rate
  reflects shared selective auditory attention.
\newblock {\em Journal of neural engineering}, 17(4):046028, 2020.

\bibitem{zhang2023multimodal}
Yukun Zhang, Shuang Qiu, and Huiguang He.
\newblock Multimodal motor imagery decoding method based on temporal spatial
  feature alignment and fusion.
\newblock {\em Journal of Neural Engineering}, 20(2):026009, 2023.

\bibitem{kleinbub2020physiological}
Johann~R Kleinbub, Alessandro Talia, and Arianna Palmieri.
\newblock Physiological synchronization in the clinical process: A research
  primer.
\newblock {\em Journal of Counseling Psychology}, 67(4):420, 2020.

\bibitem{li2021align}
Junnan Li, Ramprasaath Selvaraju, Akhilesh Gotmare, Shafiq Joty, Caiming Xiong,
  and Steven Chu~Hong Hoi.
\newblock Align before fuse: Vision and language representation learning with
  momentum distillation.
\newblock {\em Advances in neural information processing systems},
  34:9694--9705, 2021.

\bibitem{shen2022contrastive}
Xinke Shen, Xianggen Liu, Xin Hu, Dan Zhang, and Sen Song.
\newblock Contrastive learning of subject-invariant eeg representations for
  cross-subject emotion recognition.
\newblock {\em IEEE Transactions on Affective Computing}, 14(3):2496--2511,
  2022.

\bibitem{houben2015relation}
Marlies Houben, Wim Van Den~Noortgate, and Peter Kuppens.
\newblock The relation between short-term emotion dynamics and psychological
  well-being: A meta-analysis.
\newblock {\em Psychological bulletin}, 141(4):901, 2015.

\bibitem{liu2024eeg}
Huan Liu, Tianyu Lou, Yuzhe Zhang, Yixiao Wu, Yang Xiao, Christian~S Jensen,
  and Dalin Zhang.
\newblock Eeg-based multimodal emotion recognition: a machine learning
  perspective.
\newblock {\em IEEE Transactions on Instrumentation and Measurement}, 2024.

\bibitem{liu2024capsnet}
Shuaiqi Liu, Zeyao Wang, Yanling An, Bing Li, Xinrui Wang, and Yudong Zhang.
\newblock Da-capsnet: A multi-branch capsule network based on adversarial
  domain adaption for cross-subject eeg emotion recognition.
\newblock {\em Knowledge-Based Systems}, 283:111137, 2024.

\bibitem{du2020efficient}
Xiaobing Du, Cuixia Ma, Guanhua Zhang, Jinyao Li, Yu-Kun Lai, Guozhen Zhao,
  Xiaoming Deng, Yong-Jin Liu, and Hongan Wang.
\newblock An efficient lstm network for emotion recognition from multichannel
  eeg signals.
\newblock {\em IEEE Transactions on Affective Computing}, 13(3):1528--1540,
  2020.

\bibitem{yin2021eeg}
Yongqiang Yin, Xiangwei Zheng, Bin Hu, Yuang Zhang, and Xinchun Cui.
\newblock Eeg emotion recognition using fusion model of graph convolutional
  neural networks and lstm.
\newblock {\em Applied Soft Computing}, 100:106954, 2021.

\bibitem{ding2022tsception}
Yi~Ding, Neethu Robinson, Su~Zhang, Qiuhao Zeng, and Cuntai Guan.
\newblock Tsception: Capturing temporal dynamics and spatial asymmetry from eeg
  for emotion recognition.
\newblock {\em IEEE Transactions on Affective Computing}, 14(3):2238--2250,
  2022.

\bibitem{he2020advances}
Zhipeng He, Zina Li, Fuzhou Yang, Lei Wang, Jingcong Li, Chengju Zhou, and
  Jiahui Pan.
\newblock Advances in multimodal emotion recognition based on brain--computer
  interfaces.
\newblock {\em Brain sciences}, 10(10):687, 2020.

\bibitem{kwak2022fganet}
Youngchul Kwak, Woo-Jin Song, and Seong-Eun Kim.
\newblock Fganet: fnirs-guided attention network for hybrid eeg-fnirs
  brain-computer interfaces.
\newblock {\em IEEE Transactions on Neural Systems and Rehabilitation
  Engineering}, 30:329--339, 2022.

\bibitem{zitouni2022lstm}
M~Sami Zitouni, Cheul~Young Park, Uichin Lee, Leontios~J Hadjileontiadis, and
  Ahsan Khandoker.
\newblock Lstm-modeling of emotion recognition using peripheral physiological
  signals in naturalistic conversations.
\newblock {\em IEEE Journal of Biomedical and Health Informatics},
  27(2):912--923, 2022.

\bibitem{jimenez2024mmda}
Magdiel Jim{\'e}nez-Guarneros, Gibran Fuentes-Pineda, and Jonas Grande-Barreto.
\newblock Mmda: A multimodal and multisource domain adaptation method for
  cross-subject emotion recognition from eeg and eye movement signals.
\newblock {\em IEEE Transactions on Computational Social Systems}, 2024.

\bibitem{li2024incongruity}
Jing Li, Ning Chen, Hongqing Zhu, Guangqiang Li, Zhangyong Xu, and Dingxin
  Chen.
\newblock Incongruity-aware multimodal physiology signals fusion for emotion
  recognition.
\newblock {\em Information Fusion}, 105:102220, 2024.

\bibitem{chen2020simple}
Ting Chen, Simon Kornblith, Mohammad Norouzi, and Geoffrey Hinton.
\newblock A simple framework for contrastive learning of visual
  representations.
\newblock In {\em International conference on machine learning}, pages
  1597--1607. PMLR, 2020.

\bibitem{qu2020coda}
Yanru Qu, Dinghan Shen, Yelong Shen, Sandra Sajeev, Jiawei Han, and Weizhu
  Chen.
\newblock Coda: Contrast-enhanced and diversity-promoting data augmentation for
  natural language understanding.
\newblock {\em arXiv preprint arXiv:2010.08670}, 2020.

\bibitem{li2022supervised}
Yang Li, Guanyu Qiao, Xin Gao, and Guohua Wang.
\newblock Supervised graph co-contrastive learning for drug--target interaction
  prediction.
\newblock {\em Bioinformatics}, 38(10):2847--2854, 2022.

\bibitem{liu2023self}
Ziyu Liu, Azadeh Alavi, Minyi Li, and Xiang Zhang.
\newblock Self-supervised contrastive learning for medical time series: A
  systematic review.
\newblock {\em Sensors}, 23(9):4221, 2023.

\bibitem{kan2023self}
Haoning Kan, Jiale Yu, Jiajin Huang, Zihe Liu, Heqian Wang, and Haiyan Zhou.
\newblock Self-supervised group meiosis contrastive learning for eeg-based
  emotion recognition.
\newblock {\em Applied Intelligence}, 53(22):27207--27225, 2023.

\bibitem{liu2024joint}
Qile Liu, Zhihao Zhou, Jiyuan Wang, and Zhen Liang.
\newblock Joint contrastive learning with feature alignment for cross-corpus
  eeg-based emotion recognition.
\newblock {\em arXiv preprint arXiv:2404.09559}, 2024.

\bibitem{singh2023subject}
Khushboo Singh, Mitul~Kumar Ahirwal, and Manish Pandey.
\newblock Subject wise data augmentation based on balancing factor for
  quaternary emotion recognition through hybrid deep learning model.
\newblock {\em Biomedical Signal Processing and Control}, 86:105075, 2023.

\bibitem{lopez2023hypercomplex}
Eleonora Lopez, Eleonora Chiarantano, Eleonora Grassucci, and Danilo
  Comminiello.
\newblock Hypercomplex multimodal emotion recognition from eeg and peripheral
  physiological signals.
\newblock In {\em 2023 IEEE International Conference on Acoustics, Speech, and
  Signal Processing Workshops (ICASSPW)}, pages 1--5. IEEE, 2023.

\bibitem{jiang2023multimodal}
Wei-Bang Jiang, Xuan-Hao Liu, Wei-Long Zheng, and Bao-Liang Lu.
\newblock Multimodal adaptive emotion transformer with flexible modality inputs
  on a novel dataset with continuous labels.
\newblock In {\em Proceedings of the 31st ACM International Conference on
  Multimedia}, pages 5975--5984, 2023.

\bibitem{vaswani2017attention}
A~Vaswani.
\newblock Attention is all you need.
\newblock {\em Advances in Neural Information Processing Systems}, 2017.

\bibitem{darcet2023vision}
Timoth{\'e}e Darcet, Maxime Oquab, Julien Mairal, and Piotr Bojanowski.
\newblock Vision transformers need registers.
\newblock {\em arXiv preprint arXiv:2309.16588}, 2023.

\bibitem{koelstra2011deap}
Sander Koelstra, Christian Muhl, Mohammad Soleymani, Jong-Seok Lee, Ashkan
  Yazdani, Touradj Ebrahimi, Thierry Pun, Anton Nijholt, and Ioannis Patras.
\newblock Deap: A database for emotion analysis; using physiological signals.
\newblock {\em IEEE transactions on affective computing}, 3(1):18--31, 2011.

\bibitem{katsigiannis2017dreamer}
Stamos Katsigiannis and Naeem Ramzan.
\newblock Dreamer: A database for emotion recognition through eeg and ecg
  signals from wireless low-cost off-the-shelf devices.
\newblock {\em IEEE journal of biomedical and health informatics},
  22(1):98--107, 2017.

\bibitem{bradley1994measuring}
Margaret~M Bradley and Peter~J Lang.
\newblock Measuring emotion: the self-assessment manikin and the semantic
  differential.
\newblock {\em Journal of behavior therapy and experimental psychiatry},
  25(1):49--59, 1994.

\bibitem{yang2018emotion}
Yilong Yang, Qingfeng Wu, Ming Qiu, Yingdong Wang, and Xiaowei Chen.
\newblock Emotion recognition from multi-channel eeg through parallel
  convolutional recurrent neural network.
\newblock In {\em 2018 international joint conference on neural networks
  (IJCNN)}, pages 1--7. IEEE, 2018.

\bibitem{fan2024light}
Cunhang Fan, Jinqin Wang, Wei Huang, Xiaoke Yang, Guangxiong Pei, Taihao Li,
  and Zhao Lv.
\newblock Light-weight residual convolution-based capsule network for eeg
  emotion recognition.
\newblock {\em Advanced Engineering Informatics}, 61:102522, 2024.

\bibitem{li2022emotion}
Chang Li, Bin Wang, Silin Zhang, Yu~Liu, Rencheng Song, Juan Cheng, and Xun
  Chen.
\newblock Emotion recognition from eeg based on multi-task learning with
  capsule network and attention mechanism.
\newblock {\em Computers in biology and medicine}, 143:105303, 2022.

\bibitem{chen2024eeg}
Yiyuan Chen, Xiaodong Xu, Xiaoyi Bian, and Xiaowei Qin.
\newblock Eeg emotion recognition based on ordinary differential equation graph
  convolutional networks and dynamic time wrapping.
\newblock {\em Applied Soft Computing}, 152:111181, 2024.

\bibitem{sun2024emotion}
Weitong Sun and Yuping Su.
\newblock Emotion recognition of eeg based on dual-input multi-network fusion
  features.
\newblock In {\em 2024 7th International Conference on Information
  Communication and Signal Processing (ICICSP)}, pages 809--817. IEEE, 2024.

\bibitem{li2023tacoformer}
Xinda Li.
\newblock Tacoformer: Token-channel compounded cross attention for multimodal
  emotion recognition.
\newblock {\em arXiv preprint arXiv:2306.13592}, 2023.

\bibitem{li2023emotion}
Qi~Li, Yunqing Liu, Fei Yan, Qiong Zhang, and Cong Liu.
\newblock Emotion recognition based on multiple physiological signals.
\newblock {\em Biomedical Signal Processing and Control}, 85:104989, 2023.

\bibitem{sun2025msdsanet}
Weitong Sun, Xingya Yan, Yuping Su, Gaihua Wang, and Yumei Zhang.
\newblock Msdsanet: Multimodal emotion recognition based on multi-stream
  network and dual-scale attention network feature representation.
\newblock {\em Sensors (Basel, Switzerland)}, 25(7):2029, 2025.

\bibitem{liu2021comparing}
Wei Liu, Jie-Lin Qiu, Wei-Long Zheng, and Bao-Liang Lu.
\newblock Comparing recognition performance and robustness of multimodal deep
  learning models for multimodal emotion recognition.
\newblock {\em IEEE Transactions on Cognitive and Developmental Systems},
  14(2):715--729, 2021.

\bibitem{kingma2014adam}
Diederik~P Kingma.
\newblock Adam: A method for stochastic optimization.
\newblock {\em arXiv preprint arXiv:1412.6980}, 2014.

\bibitem{loshchilov2016sgdr}
Ilya Loshchilov and Frank Hutter.
\newblock Sgdr: Stochastic gradient descent with warm restarts.
\newblock {\em arXiv preprint arXiv:1608.03983}, 2016.

\bibitem{he2022adversarial}
Zhipeng He, Yongshi Zhong, and Jiahui Pan.
\newblock An adversarial discriminative temporal convolutional network for
  eeg-based cross-domain emotion recognition.
\newblock {\em Computers in biology and medicine}, 141:105048, 2022.

\bibitem{xu2025mitigation}
Zhangyong Xu, Ning Chen, Guangqiang Li, Jing Li, Hongqing Zhu, and Zhiying Zhu.
\newblock The mitigation of heterogeneity in temporal scale among different
  cortical regions for eeg emotion recognition.
\newblock {\em Knowledge-Based Systems}, 309:112826, 2025.

\bibitem{xiao2025emotionmil}
Jun Xiao, Feifei Qi, Lingli Wang, Yanbin He, Jingang Yu, Wei Wu, Zhuliang Yu,
  Yuanqing Li, Zhenghui Gu, and Tianyou Yu.
\newblock Emotionmil: An end-to-end multiple instance learning framework for
  emotion recognition from eeg signals.
\newblock {\em IEEE Transactions on Affective Computing}, 2025.

\bibitem{golland2018affect}
Yulia Golland, Adam Hakim, Tali Aloni, Stacey Schaefer, and Nava Levit-Binnun.
\newblock Affect dynamics of facial emg during continuous emotional
  experiences.
\newblock {\em Biological psychology}, 139:47--58, 2018.

\bibitem{simola2015affective}
Jaana Simola, Kevin Le~Fevre, Jari Torniainen, and Thierry Baccino.
\newblock Affective processing in natural scene viewing: Valence and arousal
  interactions in eye-fixation-related potentials.
\newblock {\em NeuroImage}, 106:21--33, 2015.

\bibitem{chanel2009short}
Guillaume Chanel, Joep~JM Kierkels, Mohammad Soleymani, and Thierry Pun.
\newblock Short-term emotion assessment in a recall paradigm.
\newblock {\em International Journal of Human-Computer Studies},
  67(8):607--627, 2009.

\bibitem{2009Feature}
David Zhang, Fengxi Song, Yong Xu, and Zhizhen Liang.
\newblock {\em Feature Level Fusion}.
\newblock Advanced Pattern Recognition Technologies with Applications to
  Biometrics, 2009.

\bibitem{zhang2009decision}
David Zhang, Fengxi Song, Yong Xu, and Zhizhen Liang.
\newblock Decision level fusion.
\newblock In {\em Advanced pattern recognition technologies with applications
  to biometrics}, pages 328--348. IGI Global, 2009.

\bibitem{xu2015adaptive}
Yong Xu and Yuwu Lu.
\newblock Adaptive weighted fusion: a novel fusion approach for image
  classification.
\newblock {\em Neurocomputing}, 168:566--574, 2015.

\bibitem{dai2021attentional}
Yimian Dai, Fabian Gieseke, Stefan Oehmcke, Yiquan Wu, and Kobus Barnard.
\newblock Attentional feature fusion.
\newblock In {\em Proceedings of the IEEE/CVF winter conference on applications
  of computer vision}, pages 3560--3569, 2021.

\bibitem{hou2019cross}
Ruibing Hou, Hong Chang, Bingpeng Ma, Shiguang Shan, and Xilin Chen.
\newblock Cross attention network for few-shot classification.
\newblock {\em Advances in neural information processing systems}, 32, 2019.

\end{thebibliography}

\end{document}